\documentclass{ieeeaccess}
\usepackage{cite}
\usepackage{amsmath,amssymb,amsfonts}
\usepackage{algorithmic}
\usepackage{graphicx}
\usepackage{textcomp}
\usepackage{lscape}
\usepackage{longtable}
\usepackage{booktabs}
\usepackage{multirow}

\usepackage{setspace}
\def\BibTeX{{\rm B\kern-.05em{\sc i\kern-.025em b}\kern-.08em
    T\kern-.1667em\lower.7ex\hbox{E}\kern-.125emX}}
\usepackage[caption=false,font=normalsize,labelfont=sf,textfont=sf]{subfig}
 \usepackage[ruled,vlined]{algorithm2e}
\usepackage{stfloats}
\usepackage{url}
\usepackage{verbatim}

\DeclareMathOperator*{\argmin}{arg\,min}

 \usepackage[]{hyperref}
\usepackage{latexsym}

\usepackage{xcolor}
\usepackage{tabularx}
\usepackage{array}
\usepackage{float}

\definecolor{mygreen}{RGB}{0,255,0}
\definecolor{myred}{RGB}{255,0,0}
\definecolor{myyellow}{RGB}{255,255,0}
\definecolor{myblue}{RGB}{0,0,255}
\definecolor{mypurple}{RGB}{100,0,100}

\newcommand{\statcirc}[1]{%
  \textcolor{#1}{\fontsize{20pt}{0}\selectfont\textbullet}%
}
\begin{document}
\history{Date of publication xxxx 00, 0000, date of current version xxxx 00, 0000.}
\doi{10.1109/ACCESS.2017.DOI}

\title{Video Anomaly Detection in 10 Years: A Survey and Outlook}
\author{\uppercase{Moshira Abdalla}\authorrefmark{1}, 
\uppercase{Sajid Javed\authorrefmark{1}, Muaz Al Radi\authorrefmark{1}, Anwaar Ulhaq \authorrefmark{2}and Naoufel Werghi \authorrefmark{1}}}
\address[1]{ Department of Computer Science, Khalifa University of Science and Technology, Abu Dhabi, UAE. }
\address[2]{A. Ulhaq is with the School of Engineering \& Technology, Central Queensland University Australia, 400 Kent Street, Sydney 2000, Australia.}



\corresp{Corresponding author: Sajid Javed (sajid.javed@ku.ac.ae ).}

\begin{abstract}
Video anomaly detection (VAD) holds immense importance
across diverse domains such as surveillance, healthcare, and environmental monitoring.  While numerous surveys focus on conventional VAD methods, they often lack depth in exploring specific approaches and emerging trends. This survey explores deep learning-based VAD, expanding beyond traditional supervised training paradigms to encompass emerging weakly supervised, self-supervised, and unsupervised approaches. 
A prominent feature of this review is the investigation of core challenges within the VAD paradigms including large-scale datasets, features extraction, learning methods, loss functions, regularization, and anomaly score prediction. 
Moreover, this review also investigates the vision language models (VLMs) as potent feature extractors for VAD. 
VLMs integrate visual data with textual descriptions or spoken language from videos, enabling a nuanced understanding of scenes crucial for anomaly detection. By addressing these challenges and proposing future research directions, this review aims to foster the development of robust and efficient VAD systems leveraging the capabilities of VLMs for enhanced anomaly detection in complex real-world scenarios. This comprehensive analysis seeks to bridge existing knowledge gaps, provide researchers with valuable insights, and contribute to shaping the future of VAD research.
\end{abstract}

\titlepgskip=-15pt

\maketitle

\section{Introduction}
\label{sec:introduction}
\PARstart{A}{nomaly}
 detection aims to pinpoint events or patterns that stray from the typical or expected behavior within a given data modality  \cite{ramachandra2020survey, aggarwal2013applications}.
It holds diverse applications, spanning from fraud detection and cybersecurity for network intrusion detection to quality assurance in manufacturing, fault identification in industrial machinery, healthcare monitoring, and beyond. Our primary emphasis, however, revolves around video anomaly detection (VAD) \cite{jiang2023weakly}, a pivotal technology that automates the identification of unusual or suspicious activities within video sequences.

While conventional methods for VAD have been extensively studied, the rapid advancements in deep learning techniques have opened up new avenues for more effective anomaly detection \cite{pang2021deep}. 
Deep learning algorithms, such as convolutional neural networks (CNNs) and vision transformers  (ViTs), have shown remarkable capabilities in learning complex patterns and representations from large-scale data \cite{shao2021transmil, transcnn}. These advancements have led to significant improvements in VAD performance, enabling more accurate and reliable detection of anomalies in video data. By leveraging deep learning techniques, researchers and practitioners have been able to develop innovative approaches that outperform traditional methods and address the limitations associated with traditional ML  \cite{nayak2021comprehensive, 2018overview}.

\color{black}
Alongside the deep learning-based VAD systems that rely on supervised learning paradigms, 
recent years have witnessed a surge in exploring novel approaches, including weakly supervised, self-supervised, and unsupervised methods \cite{georgescu2021anomaly,liu2023unsupervised, sultani2018real}.
These alternative approaches present promising solutions to the challenges encountered by conventional VAD methods, such as the requirement for extensively annotated datasets and the complexity of capturing intricate spatiotemporal patterns. Through the utilization of deep learning techniques, researchers aspire to cultivate robust and efficient VAD systems adept at navigating diverse real-world scenarios and applications \cite{nawaratne2019spatiotemporal,ramachandra2020survey}.
Similarly, several research challenges are either not thoroughly explored or remain unexplored. 
For instance, a crucial consideration is the \textit{quality and diversity of available datasets}.
The development of emerging comprehensive datasets, such as the UCF Crime dataset \cite{sultani2018real}, XD-Violence dataset \cite{xdviolence}, and ShanghaiTech dataset \cite{ShanghaiTech}, plays a pivotal role in this progress. These datasets encompass various types of anomalies, significantly contributing to the enhancement of VAD techniques.

In VAD, videos consist of sequences of frames, resulting in complex, high-dimensional spatiotemporal data. 
Detecting anomalies within such complex data necessitates methods capable of \textit{effectively capturing spatial, temporal, spatiotemporal, and textual features}.
To address these challenges, numerous VAD methods and deep feature extractors have been introduced in academic research, which has played a significant role in advancing the current state-of-the-art.
It is also worth mentioning the importance of choosing a correct loss function based on the type of VAD paradigm.
Hence, \textit{loss functions and regularization} techniques are another challenge in the VAD problem.
Another significant challenge lies in the diverse paradigms including \textit{self-supervised, weakly supervised, fully supervised, and unsupervised models} utilized to address the VAD problem.

These paradigms can be categorized into classical and deep learning. 
The classical approaches employ handcrafted features including spatiotemporal gradient \cite{lu2013abnormal}, Histograms of Oriented Gradients (HOG) features \cite{hasan2016learning, HOG}, and Histograms of Optical Flows (HOF) \cite{HOF} within a temporal cuboid.
These features are selected for their effectiveness in capturing appearance and motion information in a spatiotemporal context.

The deep learning approaches are more powerful in terms of rich feature representation and end-to-end learning and have gained much popularity in the past decade due to advancements in different learning models. 
Specifically, these approaches leverage powerful feature extractors to capture meaningful spatiotemporal features. 
Examples include convolutional neural networks \cite{luo2017remembering}, autoencoders \cite{liu2021hybrid}, GANs \cite{li2023multi}, vision transformers \cite{sun2023hierarchical,li2022self}, and vision language models \cite{ni2022expanding, ju2022prompting}.

In addition to the aforementioned inherent challenges within the VAD problem, each paradigm comes up with diverse \textit{loss functions, spatiotemporal regularization/constraints, and anomaly score prediction components}.
We also identify these different modules in this survey and provide take-home messages for each of the abovementioned challenges.
 Figure ~\ref{fig:introchart} shows the performance variation of representative deep learning-based VAD methods on two publicly available benchmark datasets including UCF-Crime \cite{sultani2018real} and ShanghaiTech \cite{ShanghaiTech}. 
The performance is reported in terms of the area under the ROC curve (AUC\%). 
The performance improvement trend in this figure underscores a significant advancement in deep learning methodologies over the past decade, with the most notable improvement achieved by utilizing the recently proposed vision-language-based model \cite{CLIPTSA}.

\color{black}
    \Figure[t!](topskip=0pt, botskip=0pt, midskip=0pt)[width=0.999\linewidth]{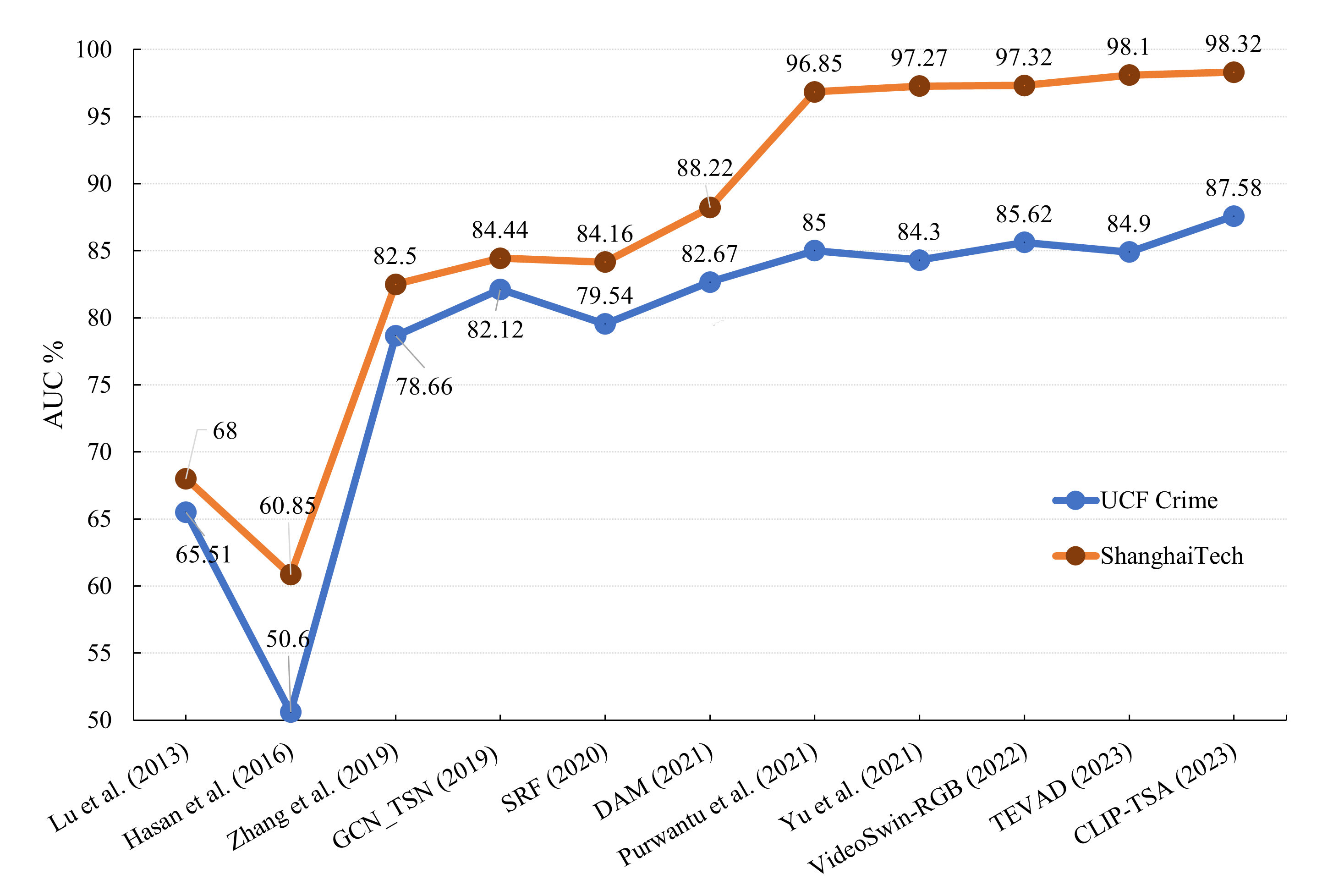}{Performance improvement from 2017 until 2023 on two popular benchmarks. Performance is measured by the area under the ROC curve (AUC\%). Note that some of the models were developed before the datasets were created but were used after the creation by other researchers such as \cite{lu2013abnormal,hasan2016learning},  the other proposed models are: \cite{zhang2019temporal},\cite{zhong2019graph},\cite{ zaheer2020self},\cite{DAM},\cite{purwanto2021dance},\cite{tian2021weakly},\cite{li2022self},\cite{TEVAD},\cite{CLIPTSA}.\label{fig:intro chart}}

\textbf{Motivation}: Despite the growing interest in VAD and the proliferation of deep learning-based approaches, there remains a need for a comprehensive survey that explores the latest developments in this field. Existing surveys often focus on traditional VAD methods and may not adequately cover the emerging trends and methodologies such as vision-language models (VLMs) \cite{CLIP,wu2023vadclip,TEVAD,lin2022swinbert,CLIPTSA}.
By delving into VLMs and deep learning-based VAD, including supervised, weakly supervised, self-supervised, and unsupervised approaches, this survey aims to provide a thorough understanding of the state-of-the-art techniques and their potential applications.
Our main focus lies on deep learning-based solutions for VAD problems.
Particularly, we explore emerging VAD paradigms that employ different datasets, feature extractions, loss functions, and spatiotemporal regularization. 
In addition, we offer a comprehensive analysis of over 50 different methodologies employed in VAD, with a specific focus on the potential of textual features extracted 
with vision language models. 
In this survey, our main contributions are summarized as follows:

\begin{itemize}

\item We identify the core challenges of the SOTA VAD paradigms including the vision-language models in terms of feature extraction, large-scale VAD training datasets, loss functions, spatiotemporal regularization, and video anomaly scores prediction. 
\item We provide a comparative analysis through quantitative and qualitative comparisons of SOTA models on different benchmarking datasets. 
This analysis sheds light on the strengths and weaknesses of existing methodologies, offering valuable insights for researchers and practitioners in the field.
\item Following our analysis, we outline our proposed recommendations for addressing the open challenges in VAD. These recommendations are informed by our deep understanding of the current landscape of VAD research and aim to guide future research directions toward overcoming existing limitations and advancing the SOTA.

\end{itemize}

The structure of this paper is as follows: Section \ref{review method} outlines the methodology used to select the research studies included in this review over the past decade. Section \ref{Literature} examines previous surveys conducted in the field of video anomaly detection. Section \ref{problem} presents the formulation of the video anomaly detection problem. Section \ref{challenges} presents a systematic approach and taxonomy for analyzing the VAD problem, and then the core challenges including the datasets \ref{datasets}, feature extraction \ref{feature extraction}, supervision schemes \ref{DL methods}, loss functions \ref{loss func}, regularization techniques \ref{reg} and the anomaly score \ref{anomaly score}.
Section \ref{expermentation and evaluation} outlines datasets guidelines and evaluation protocols used. Section \ref{comparative analysis} provides a comparative analysis through quantitative and qualitative comparisons of state-of-the-art models. Section \ref{Bibliometric} provides visualization of bibliometric networks for thematic analysis. Finally, we conclude our work and present additional future directions in Section \ref{conc}


\section{Review methodology} \label{review method}

This survey paper exclusively considers research directly related to ``Video Anomaly Detection." 
The survey is conducted systematically, utilizing publications from top computer vision venues, including CVPR (Conference on Computer Vision and Pattern Recognition), ICCV (International Conference on Computer Vision), ECCV (European Conference on Computer Vision), IEEE TPAMI (Transactions on Pattern Analysis and Machine Intelligence), IJCV (International Journal of Computer Vision), CVIU (Computer Vision and Image Understanding).
The objective of this review paper is to present the state-of-the-art in Video Anomaly Detection. 
Accordingly, this study focuses on identifying published research concerning the implementation of various computer Vision and deep Learning-based methods to address the challenges of VAD. 
The reviewed works span the last decade covering over 50 articles. 


\section{Related work} \label{Literature}
In the field of anomaly detection, several survey papers have been conducted in the past decade \cite{2018overview, chalapathy2019deep, ramachandra2020survey, nayak2021comprehensive, pang2021deep}. The first survey paper, published in 2018 by \cite{2018overview}, focused on deep learning techniques for VAD, with particular attention to unsupervised and semi-supervised methods. Their classification of models included three distinct categories: reconstruction-based, spatio-temporal predictive, and generative models. Notably, this paper preceded the notable Multiple Instance Learning (MIL) approach, resulting in the omission of weakly supervised methods from their study.

Another significant study by Chalapathy \textit{et al.} \cite{chalapathy2019deep} highlighted the potential of deep anomaly detection methods in addressing various detection challenges. Their work covered diverse application areas such as the Internet of Things, intrusion detection, and surveillance videos. They further classified anomalies into collective, contextual, and point anomalies. The deep learning methods they categorized into four primary classes include unsupervised, semi-supervised, hybrid, and One-Class Neural Network models.

In another survey, Ramachandra \textit{et al.} \cite{ramachandra2020survey} focus primarily on detecting anomalies within a single scene while also highlighting the differences from multi-scene anomaly detection. An important distinction lies in the fact that single-scene VAD  may involve anomalies dependent on specific locations, whereas multi-scene detection cannot. The survey also sheds light on benchmark datasets employed for single-scene versus multi-scene detection and the associated evaluation procedures. In a broader context, the survey classifies previous research in video anomaly detection into three main categories: distance-based, probabilistic, and reconstruction-based approaches.

In a different work, Nayak \textit{et al.} \cite{nayak2021comprehensive} categorize learning frameworks into four main categories: supervised, unsupervised, semi-supervised, and active learning. In the context of deep-learning-based VAD, state-of-the-art methods fall into several distinct categories, including Trajectory-based methods, Global pattern-based methods, Grid pattern-based methods, Representation learning models, Discriminative models, Predictive models, Deep generative models, Deep one-class deep neural networks, and Deep hybrid models. Moreover, the research provides a comprehensive analysis of performance evaluation methodologies, covering aspects such as the choice of datasets, computational infrastructure, evaluation criteria, and performance metrics. 

Another work by Pang \textit{et al.} \cite{pang2021deep} focused on deep learning techniques for anomaly detection and explored various challenges within the anomaly detection (AD) problem. These challenges included issues such as class imbalance in the data, complex anomaly detection, the presence of noisy instances in weakly supervised AD methods, and more. They discussed how deep learning methods offer solutions to these diverse challenges. To structure their analysis, Pang \textit{et al.} introduced a hierarchical taxonomy for categorizing deep anomaly detection methods. This taxonomy comprised three principal categories: deep learning for feature extraction, learning feature representations of normality, and end-to-end anomaly score learning. Additionally, they provided 11 fine-grained subcategories from a modeling perspective.

In their research, Mohammad Baradaran and Robert Bergevin \cite{baradaran2024critical} delve deeply into the semi-supervised   VAD approaches, focusing on scenarios where labeled anomaly data is limited. They emphasize the role of feature extractors in these contexts, highlighting their ability to differentiate intricate patterns within video data by capturing crucial spatial and temporal details. These feature extractors are pivotal for detecting anomalies in semi-supervised tasks, where the model learns predominantly from an extensive set of normal data. The authors conduct an experimental analysis to shed light on the strengths and weaknesses of various VAD methods. They categorize the DL semi-supervised approaches into six distinct types: reconstruction, prediction, memorization, object-centric,  segmentation, and multi-task learning-based methods. For each category, they provide a comprehensive examination of the strengths and shortcomings, particularly focusing on the effectiveness of different feature extraction techniques.

Another recently published survey paper by Nomica \textit{et al.} \cite{choudhry2023comprehensive} provides an in-depth analysis of machine learning techniques for detecting anomalies in video surveillance systems. It categorizes these methods into supervised, semi-supervised, and unsupervised approaches, highlighting their strengths, weaknesses, and applicability. However, it does not address the critical differences between feature types—such as temporal, spatial, textual, and hybrid features. These differences significantly influence the choice of feature extractors, impacting the effectiveness of the detection models. Additionally, it significantly overlooks the topic of  Vision-Language Models as feature extractors. In contrast, our work provides an in-depth analysis that encompasses all these aspects, offering a comprehensive understanding of their impact on anomaly detection models.

A very recently published survey paper in 2024 by Yang  \textit{et al.} \cite{liu2023generalized} categorizes the VAD approaches into unsupervised, Weakly-supervised, fully unsupervised, and supervised VAD. It highlights their strengths, such as improved feature extraction and detailed object analysis, while also noting the importance of handling spatiotemporal features and illumination changes. Although the survey provides an extensive comparison of different VAD methodologies, it notably omits the discussion of Vision-Language Models as feature extractors, a growing area of interest in the field.

\color{black}

Previous research efforts have overlooked the critical importance of employing diverse feature extractors within deep learning models. This oversight becomes apparent when considering emerging trends such as the widespread adoption of transformers and vision-language pretraining models within video anomaly detection. These innovations have significantly influenced the overall performance of deep learning models. In our survey paper, we aim to address these research gaps that have not been adequately outlined in previous surveys.

Additionally, we classify learning and supervision methods into four distinct groups: supervised, unsupervised, self-supervised, and weakly supervised techniques. This classification scheme allows us to systematically analyze and compare different approaches, providing readers with valuable insights into the diverse methodologies employed in video anomaly detection research. By organizing our study in this manner, we aim to facilitate a deeper understanding of the underlying principles and techniques driving advancements in this field.

\subsection{Contributions}

In this survey, our main contributions compared to existing survey papers are summarized as follows:

\begin{itemize}
    \item This work defines a clear and comprehensive problem formulation of the VAD problem specifically in the context of supervised learning, where frame-level labels are available. 
    
    \item To the best of our knowledge, this is the first survey to highlight the emerging importance of integrating Vision-Language Models (VLMs) as feature extraction in VAD. We explore how VLMs can significantly enhance model performance by effectively combining visual and textual data to better understand and detect anomalies.
    
    \item  The paper is organized to serve as a detailed guide for readers and researchers new to the field of VAD. We provide foundational knowledge and a structured approach to navigate the complexities of VAD research as shown in Figure \ref{fig:anomaly formulation.png}.
    \item This work presents a well-structured discussion starting from
the selection of datasets and the types of features that
are critical for VAD. This includes a focus on textual
features and deep feature extractors. We also explore
the various learning and supervision paradigms, including supervised, self-supervised, weakly supervised, and
unsupervised/reconstruction approaches, detailing their
respective advantages and disadvantages.
    \item   This work provides a taxonomy of video anomaly detection that is systematically categorized into two main dimensions: learning and supervision schemes, and feature extraction as shown in Figure \ref{Taxonamy}

    \item This work provides insights into selecting appropriate loss functions and regularization techniques, which are crucial for optimizing the performance of VAD models.
    
    \item  A comprehensive guideline is provided for choosing the most suitable datasets and evaluation metrics for experimental purposes, ensuring that researchers can effectively assess and compare their VAD methods.
    
    \item This work offers a detailed comparative analysis of the state-of-the-art (SOTA) VAD methods, both quantitatively and qualitatively. This analysis helps in understanding the current landscape and performance benchmarks in the field.
    
    \item  Finally, this work provides an extensive discussion on potential future research directions in VAD. This includes exploring new technologies, methodologies, and application areas, and guiding researchers toward promising avenues for further investigation.
\end{itemize}


\color{black}

\section{Defining the Video Anomaly Detection Problem} \label{problem} 
Anomalies within video data encompass events or behaviors that notably diverge from anticipated or regular patterns. The primary goal of video anomaly detection is to devise and implement resilient algorithms and models capable of autonomously identifying and flagging these anomalies in real-time. This involves converting raw video data into interpretable feature representations utilizing robust feature extractors adept at capturing both spatial and temporal characteristics. Additionally, it necessitates the selection of appropriate algorithms or techniques and the establishment of effective evaluation metrics to assess detection performance accurately.

In the context of supervised learning scenarios where frame-level labels are available, video anomaly detection problem can be succinctly described as follows:

\noindent We represent each video as $V_i$, which consists of a sequence of frames $\{f_{i,1}, f_{i,2}, \dots, f_{i,n}\}$. From each frame, we can extract essential feature representations, denoted by $x_{i,j}$. Define a model $M$ that takes these features extracted from each frame and produces an anomaly score for that frame. For each frame $f_{i,j}$, the anomaly score is $S(f_{i,j}) = M(x_{i,j})$. 

\noindent The total anomaly score for video $V_i$ is the sum of the scores of its frames:
\begin{equation}
    S(V_i) = \sum_{j=1}^{n} S(f_{i,j}) 
\end{equation}


\noindent This anomaly score $S(V_i)$ is compared with a predetermined threshold $T$, we can define the predicted binary label, $\hat{Y_i}$, for the video as:

\[
\hat{Y_i} = 
\begin{cases} 
1 & \text{if } S(V_i) \geq T \\
0 & \text{if } S(V_i) < T 
\end{cases}
\]

\noindent Where: $1$ indicates that $V_i$ is anomalous and $0$ indicates that $V_i$ is normal.  The true label of the video is represented as $Y_i \in \{0, 1\}$. The objective is to train the model $M$ such that the difference between $\hat{Y_i}$ and $Y_i$ is minimized for all videos in the training set and generalizes well to unseen videos.

\section{Analyzing Video Anomaly Detection: A Systematic Approach} \label{challenges}

\begin{figure*}[h!]
\centering
\includegraphics[width=\linewidth]{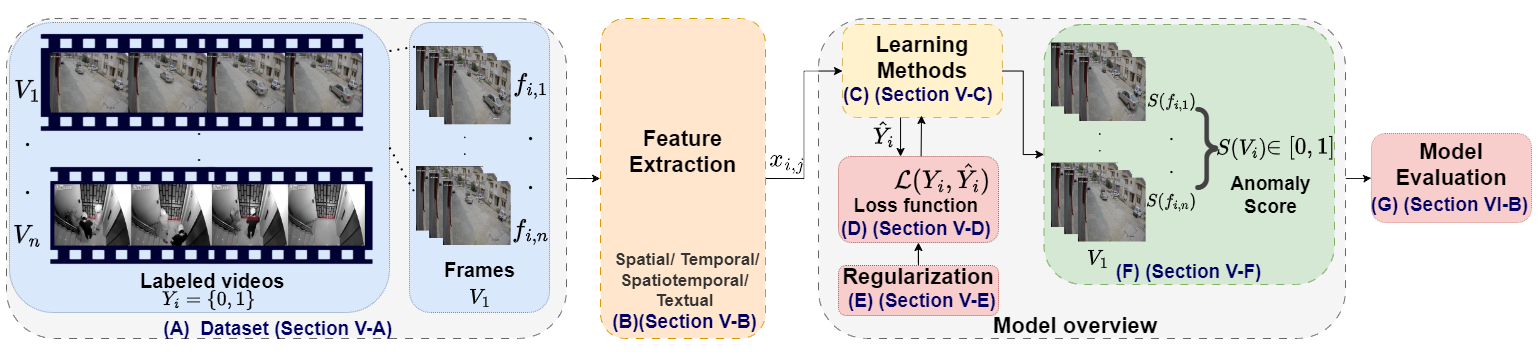}

\caption{ \small 
Video anomaly detection paradigm including (A) state-of-the-art dataset building and selection \ref{datasets}, (B) Spatial, temporal, spatio-temporal, and textual deep feature extraction \ref{feature extraction}, (C) Diverse deep learning and supervision schemes (supervised, self-supervised, weakly supervised, and unsupervised methods) \ref{DL methods}, (D) selection of loss functions \ref{loss func}, (E) integration of regularization techniques within loss functions \ref{reg}, (F) anomaly score calculation \ref{anomaly score}, and (G) model evaluation techniques \ref{evaluation}. }

\centering
\label{fig:anomaly formulation.png}
\end{figure*}

In this section, we dig deeper into the complexities of the VAD process by analyzing relevant literature. The diagram presented in Figure \ref{fig:anomaly formulation.png} acts as a guide for navigating VAD research in the subsequent sections, where we explore the challenges associated with the VAD problem. Commencing with an array of diverse datasets, we traverse through various feature extraction techniques utilized to extract spatial, temporal, spatiotemporal, or textual features, leveraging vision language models.

Our exploration begins with the utilization of diverse datasets and a spectrum of feature extraction techniques. These methodologies are applied to extract spatial, temporal, spatiotemporal, and textual features.
Furthermore, our exploration will encompass various learning and supervision strategies, including supervised methods, self-supervised techniques, and unsupervised techniques (often classified as reconstruction-based or one-class classification approaches), alongside weakly supervised and prediction methods. Additionally, we will illuminate the significance of loss functions, regularization techniques, and anomaly score computation. Evaluation protocols for models are discussed in section \ref{evaluation}. In the upcoming sections, we will explore these challenges further, examining the ways they have been approached.

\paragraph{\textbf{Taxonomy of Video Anomaly Detection}}

\begin{figure*}[h!]
\centering
\includegraphics[width=\linewidth]{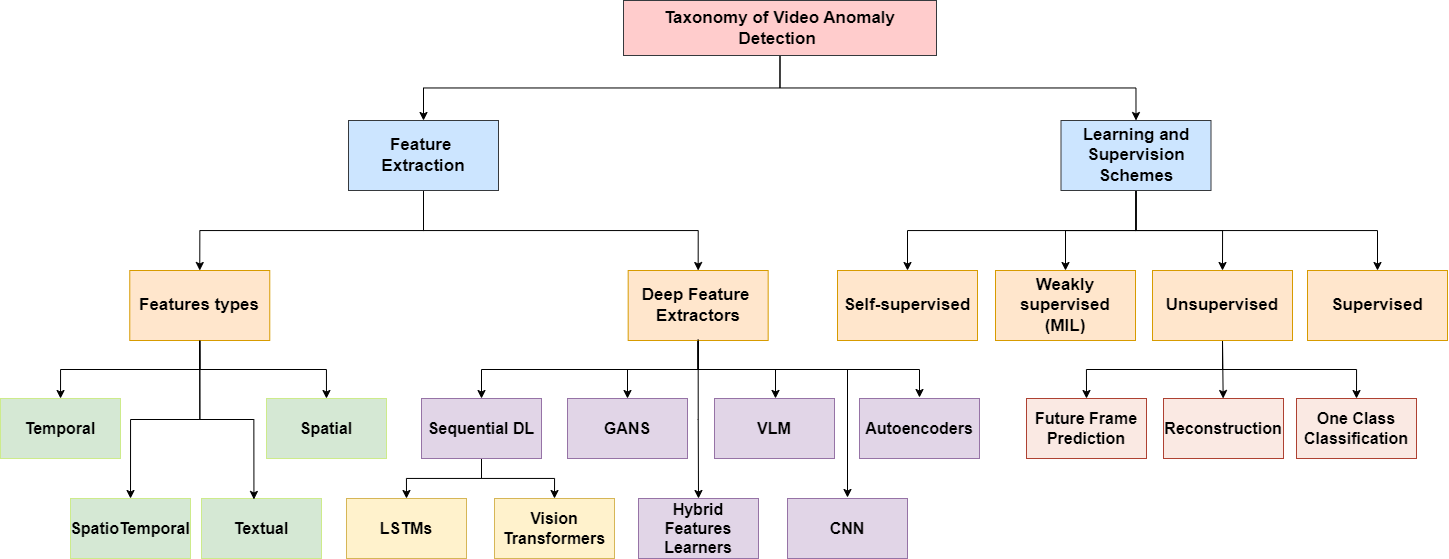}

\caption{\small Taxonomy of Video Anomaly Detection.}


\centering
\label{Taxonamy}
\end{figure*}

The taxonomy of video anomaly detection, as shown in Figure \ref{Taxonamy}, is systematically categorized into two main dimensions: Learning and Supervision Schemes, and Feature Extraction. The Learning and Supervision Schemes dimension includes Supervised, Self-Supervised, Weakly Supervised (such as Multiple Instance Learning), and Unsupervised methods. Unsupervised methods are further divided into   One-Class Classification, Reconstruction, and Future Frame Prediction approaches. The Feature Extraction dimension involves Deep Feature Extractors, including CNNs, Autoencoders, GANs, Sequential Deep Learning models (like LSTMs and Vision Transformers), Vision Language models, and Hybrid models. Additionally, it covers different types of features such as Spatial, Temporal, SpatioTemporal, and Textual.

\color{black}

\subsection{Datasets Building and Selection} \label{datasets}

The field of Video Anomaly Detection relies significantly on publicly available datasets which are used for testing and benchmarking the proposed models. In this section, we present an overview of the common datasets utilized in the field of VAD, each carefully curated to facilitate the study of anomalies in various scenarios. These datasets encompass a wide range of scenes, offering diverse challenges for anomaly detection. We will investigate the number of videos contained in each dataset, the specific types of scenes they cover, and the anomalies present.

\begin{figure*}[t] 
\centering
\frame{\includegraphics[width=\linewidth]{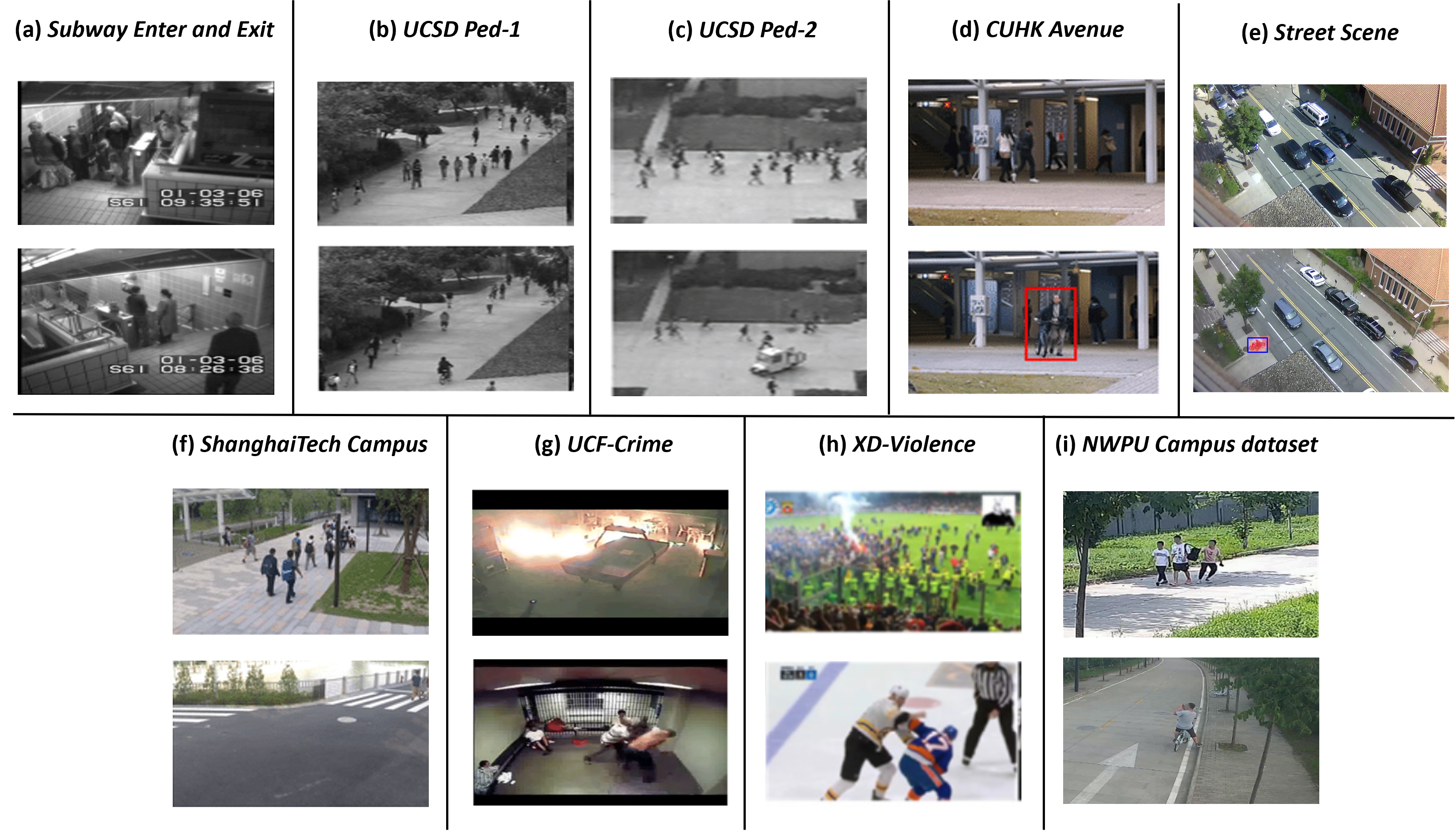}}

\caption{ \small 
 Sample frames showcasing the diversity of scenes and anomalies present in publicly available datasets used for Video Anomaly Detection. These frames offer a glimpse into the range of challenges and scenarios addressed within the field, providing valuable insights for testing and benchmarking anomaly detection models.}

\centering
\label{fig:SampleFrames}
\end{figure*}

\subsubsection{Subway dataset}  

The Subway dataset \cite{Subway} consists of two videos collected using a CCTV camera, each capturing distinct perspectives of an underground train station. The first video focuses on the ``entrance gate" area, where individuals typically descend through the turnstiles and enter the platform with their backs turned to the camera. In contrast, the second video is situated at the ``exit gate," observing the platform where passengers ascend while facing the camera. These two cameras provide unique vantage points within the station, offering valuable insights for analysis and surveillance. The dataset has a combined duration of 2 hours. Anomalies within this dataset encompass activities such as walking in incorrect directions and loitering. Notably, this dataset is recorded within an indoor environment.


 \subsubsection{UCSD Pedestrian }

The UCSD anomaly detection dataset \cite{UCSD} was collected using a stationary camera positioned at an elevated vantage point to monitor pedestrian walkways. This dataset offers a wide range of scenes depicting varying crowd densities, spanning from sparsely populated to highly congested environments. Normal videos within the dataset predominantly feature pedestrians, while abnormal events arise from two primary sources: the presence of non-pedestrian entities in the walkways and the occurrence of anomalous pedestrian motion patterns. Common anomalies observed include bikers, skaters, small carts, individuals walking across designated walkways or in adjacent grass areas, as well as instances of people in wheelchairs. The dataset comprises two distinct subsets, Peds1 and Peds2, each capturing different scenes. Peds1 depicts groups of people walking towards and away from the camera, often exhibiting perspective distortion, while Peds2 focuses on scenes where pedestrians move parallel to the camera plane.

The dataset is further divided into clips, with each clip containing approximately 200 frames. Ground truth annotations are provided at a frame level, indicating the presence or absence of anomalies. Additionally, a subset of clips in both Peds1 and Peds2 is accompanied by manually generated pixel-level binary masks, enabling the evaluation of algorithms' ability to localize anomalies.The UCSD Ped1 \& Ped2 datasets are publicly available: \hyperlink{link1}{http://www.svcl.ucsd.edu/projects/anomaly/}.\\

{

\subsubsection{Street Scene dataset} 

The Street Scene dataset proposed by Ramachandra \textit{et al.} \cite{ramachandra2020street} is designed for video anomaly detection and comprises 46 training and 35 testing high-resolution (1280×720) video sequences. These sequences were captured using a USB camera positioned to overlook a two-lane street with bike lanes and pedestrian sidewalks during the daytime. This dataset presents a challenging environment due to the diverse range of activities captured, including cars driving, turning, stopping, and parking; pedestrians walking, jogging, and pushing strollers; and bikers riding in bike lanes. Additionally, the videos feature changing shadows, moving backgrounds such as flags and trees blowing in the wind, and occlusions from trees and large vehicles.

The dataset includes 56,847 frames for training and 146,410 frames for testing, extracted at a rate of 15 frames per second. It contains a total of 205 naturally occurring anomalous events. These anomalies range from illegal activities like jaywalking and illegal U-turns to uncommon events not present in the training set, such as pets being walked and a meter maid issuing tickets. This variety makes the Street Scene dataset a comprehensive and demanding resource for advancing video anomaly detection research. However, a notable challenge of this dataset is that it is a single-scene dataset where only the street scene is included. This prevents models trained only on this dataset from generalizing their anomaly detection capabilities to other scenes. On the other hand, this dataset could be beneficial for models that are specifically designed to work on single scenes such as EVAL \cite{singh2023eval}. The dataset is publicly available on: \hyperlink{linkN}{http://www.merl.com/demos/video-anomaly-detection} 

}

 \subsubsection{UCF-Crime dataset  }

The UCF-Crime dataset \cite{sultani2018real} stands as a widely utilized large-scale dataset within recent research endeavors, boasting a multi-scene approach. Comprised of long, untrimmed surveillance videos, this dataset encapsulates 13 real-world anomalies with profound implications for public safety. The anomalies span a broad spectrum, including Abuse, Arrest, Arson, Assault, Road Accident, Burglary, Explosion, Fighting, Robbery, Shooting, Stealing, Shoplifting, and Vandalism.

To ensure the dataset's integrity and quality, a meticulous curation process was undertaken. Ten annotators were trained to collect videos from platforms like YouTube and LiveLeak, utilizing text search queries across multiple languages. Pruning conditions were enforced to exclude manually edited videos, prank videos, those captured by handheld cameras, and those sourced from news or compilations. The resultant dataset comprises 950 unedited real-world surveillance videos, each featuring clear anomalies, alongside an equal number of normal videos, totaling 1900 videos.

Temporal annotations were meticulously acquired by assigning the same videos to multiple annotators and averaging their annotations. The dataset is thoughtfully partitioned into a training set, consisting of 800 normal and 810 anomalous videos, and a testing set, comprising 150 normal and 140 anomalous videos. With its expansive coverage of various anomalous events, the UCF-Crime dataset serves as a comprehensive resource for evaluating anomaly detection algorithms across diverse real-world scenarios. The dataset is publicly available: \hyperlink{link1}{http://crcv.ucf.edu/projects/real-world/}.\\

\subsubsection{CUHK Avenue } 

The CUHK Avenue dataset \cite{lu2013abnormal} is a widely employed resource for Video Anomaly Detection, meticulously crafted by researchers from the Chinese University of Hong Kong (CUHK). Set within the CUHK campus avenue, this dataset primarily focuses on anomalies commonly encountered in urban environments and public streets. It offers a diverse array of lighting conditions, weather variations, and human activities, thereby presenting formidable challenges for VAD models.

An array of anomalous behaviors is captured within the dataset, encompassing both physical anomalies such as fighting and sudden running, and non-physical anomalies like unusual gatherings and incorrect movement directions. Notably, the dataset introduces several key challenges for VAD models, including slight camera shake in certain frames, occasional absence of normal behaviors in the training set, and outliers in the training data. In total, the dataset comprises 30,652 frames, with 15,328 frames allocated for training purposes and the remaining 15,324 frames earmarked for testing. With its diverse scenarios and realistic challenges, the CUHK Avenue dataset serves as a valuable benchmark for evaluating and advancing VAD techniques. The dataset is publicly available: \hyperlink{link1}{http://www.cse.cuhk.edu.hk/leojia/projects/detectabnormal
/dataset.html}.\\


\subsubsection{ShanghaiTech}

The ShanghaiTech Campus dataset \cite{ShanghaiTech} stands as a substantial contribution to the field of anomaly detection, presenting a vast and diverse collection of data. Across its 13 scenes, characterized by intricate light conditions and varied camera angles, the dataset encapsulates a comprehensive array of challenging scenarios. Notably, it boasts an extensive compilation of over 270,000 training frames, alongside 130 abnormal event occurrences meticulously annotated at the pixel level, facilitating precise evaluation and analysis.

Comprising a total of 330 training videos and 107 testing videos, all rendered at a resolution of 856x480, the dataset ensures consistency and compatibility across its contents. Each video maintains a frame rate of 24 frames per second (24fps), ensuring smooth and standardized playback. With its wealth of data and meticulous annotations, the ShanghaiTech Campus dataset serves as a cornerstone resource for advancing anomaly detection methodologies in real-world scenarios. The dataset is publicly available: \hyperlink{link1}{\url{https://sviplab.github.io/dataset/campus_dataset.html}}\\


\subsubsection{XD-Violence  }

The XD-Violence dataset \cite{xdviolence}  is a large-scale and multi-scene dataset with a total duration of 217 hours, containing 4754 untrimmed videos. This dataset comprises 2405 violent videos and 2349 non-violent videos, all of which include audio signals and weak labels. The focus of the dataset was on the field of weakly supervised violence detection, where only video-level labels are available in the training set. This approach offers the advantage of being more labor-saving compared to annotating frame-level labels. As a result, forming large-scale datasets of untrimmed videos and training a data-driven and practical system is no longer a challenging endeavor. The dataset incorporates audio signals and is sourced from both movies and in-the-wild scenarios. The dataset encompasses six physically violent classes: Abuse, Car Accident, Explosion, Fighting, Riot, and Shooting. The dataset is divided into a training set, which comprises 3954 videos, and a test set which includes 800 videos. Within the test set, there are 500 violent videos and 300 non-violent videos. \hyperlink{link1}{\url{https://roc-ng.github.io/XD-Violence/}}

\subsubsection{NWPU Campus dataset} 
The NWPU Campus dataset \cite{NWPU} is newly available. It represents a significant contribution to the field of video anomaly detection and anticipation. The dataset was compiled by setting up cameras at 43 outdoor locations on the campus to capture the activities of pedestrians and vehicles. To ensure a sufficient number of anomalous events, more than 30 volunteers were involved in performing both normal and abnormal activities. The dataset covers a wide range of normal events, including regular walking, cycling, driving, and other daily behaviors that adhere to rules. The anomalies encompass various categories such as single-person anomalies, interaction anomalies, group anomalies, scene-dependent anomalies, location anomalies, appearance anomalies, and trajectory anomalies. The dataset consists of 305 training videos and 242 testing videos, totaling 16 hours of footage. Frame-level annotations are provided for the testing videos, indicating the presence or absence of anomalous events. Privacy concerns are addressed by blurring the faces of volunteers and pedestrians. Compared to other datasets, the NWPU Campus dataset stands out due to its larger volume of data, diverse scenes, and inclusion of scene-dependent anomalies. Additionally, it is the first dataset designed for video anomaly anticipation. These features make the NWPU Campus dataset a valuable resource for advancing research in video anomaly detection and anticipation.    The dataset is publicly available: \hyperlink{link1}{https://campusvad.github.io/}\\

\begin{table*}[]
\centering

\caption{Overview of the most important publicly available datasets.}
\label{tab:datasets}
\resizebox{\textwidth}{!}{%
\begin{tabular}{|c|l|c|c|c|c|c|c|c|c|}
\hline
\textbf{\textbf{\textbf{Dataset Name}}} & \textbf{\textbf{\textbf{Paper}}} & \textbf{\textbf{\textbf{Year}}} & \textbf{\textbf{\textbf{Balanced}}} & \textbf{\textbf{\textbf{\# Videos}}} & \textbf{\textbf{\textbf{Length}}} & \textbf{\textbf{\textbf{Example anomalies}}}                                                                            & \textbf{\textbf{\textbf{Frames Per. Video}}} & \textbf{\textbf{\textbf{Train:Test}}} & \textbf{\textbf{Challenges}}                                                                                            \\ \hline
Subway Entrance Gate          & \cite{Subway}                   & 2008                            & No                                  & 1                                    & 1.5 hrs                             & Wrong direction, No payment                                                                                             & 121,749                                     & 13:87                                 & \begin{tabular}[c]{@{}c@{}}Single scene, indoor only,\\  limited number of anomalies\end{tabular}                       \\ \hline
Subway Exit Gate          & \cite{Subway}                   & 2008                            & No                                  & 1                                   & 1.5 hrs                             & Wrong direction, No payment                                                                                             & 64,901                                     & 13:87                                 & \begin{tabular}[c]{@{}c@{}}Single scene, indoor only,\\  limited number of anomalies\end{tabular}                       \\ \hline
UCSD Ped1                               & \cite{UCSD}                     & 2010                            & No                                  & 70                                   & 5 min                             & \begin{tabular}[c]{@{}c@{}}Bikers, small carts, \\ walking across walkways,\\  wheelchair users\end{tabular}            & 201                                       & 49:51                                 & \begin{tabular}[c]{@{}c@{}}Small size, single scene,\\  only outdoor, only\\  vehicle anomalies\end{tabular}            \\ \hline
UCSD Ped2                               & \cite{UCSD}                     & 2010                            & No                                  & 28                                   & 5 min                             & \begin{tabular}[c]{@{}c@{}}Bikers, small carts,\\  walking across walkways,\\ wheelchair users\end{tabular}             & 163                                       & 55:45                                 & \begin{tabular}[c]{@{}c@{}}Small size, single scene,\\  only outdoor, only\\  vehicle anomalies\end{tabular}            \\ \hline
CUHK Avenue                             & \cite{lu2013abnormal}                     & 2013                            & No                                  & 37                                   & 30 min                            & \begin{tabular}[c]{@{}c@{}}Strange action, \\ Wrong direction,\\  Abnormal object\end{tabular}                          & 839                                       & 50:50                                 & \begin{tabular}[c]{@{}c@{}}Small size, single scene,\\   outdoor only, camera shake\end{tabular}                                       \\ \hline
ShanghaiTech Campus Dataset             & \cite{ShanghaiTech}             & 2017                            & No                                  & 437                                  & 3.67 hrs                          & \begin{tabular}[c]{@{}c@{}}Running, Loitering,\\  Biking in Restricted Areas,\\  Unusual Gatherings, Theft\end{tabular} & 1,040                                      & 86:14                                 & \begin{tabular}[c]{@{}c@{}}Only university setting\\  anomalies, single \\ geographic location\end{tabular}             \\ \hline
UCF-Crime                               & \cite{sultani2018real}                 & 2018                            & Yes                                 & 1,900                                 & 128 hrs                           & \begin{tabular}[c]{@{}c@{}}Abuse, arrest, arson,\\  assault, accident, burglary,\\  fighting, robbery\end{tabular}      & 7,247                                      & 85:15                                 & \begin{tabular}[c]{@{}c@{}}Imbalance between normal\\  and abnormal classes, \\ Variation in Video Quality\end{tabular} \\ \hline
{Street Scene}                               & \cite{ramachandra2020street}                 & 2020                            & No                                 & 81                                 & 226 mins                           & \begin{tabular}[c]{@{}c@{}}jaywalking,\\  illegal U-turns, pets,\\   metermaid ticketing a car\end{tabular}      & 2,509                                      & 57:43                                 & \begin{tabular}[c]{@{}c@{}} Single
geographic location, \\single scene,\\  only outdoor,\end{tabular} \\ \hline
XD-Violence                             & \cite{xdviolence}               & 2020                            & No                                  & 4,754                                 & 214 hrs                           & \begin{tabular}[c]{@{}c@{}}Abuse, Car Accident,\\  Explosion, Fighting, \\ Riot, Shooting\end{tabular}                  & 3,944                                      & 83:17                                 & \begin{tabular}[c]{@{}c@{}}Limited number of\\  anomalies, Variation \\ in Video Quality\end{tabular}                   \\ \hline
NWPU Campus dataset                     & \cite{NWPU}                     & 2023                            & No                                  & 547                                  & 16 hrs                            & \begin{tabular}[c]{@{}c@{}}Single-person, interaction,\\  group, location,\\  appearance, trajectory\end{tabular}       & 2,527                                      & 56:44                                 & \begin{tabular}[c]{@{}c@{}}Only university setting\\  anomalies, single \\ geographic location\end{tabular}             \\ \hline
\end{tabular}%
}

\end{table*}

\subsubsection{Discussion on Datasets}

The existing literature presents a plethora of diverse and extensive datasets covering a broad spectrum of normal and abnormal scenarios. These datasets vary from single-scene datasets, which focus on specific situations and their anomalies, such as the UCSD Pedestrian dataset \cite{UCSD}, to more diverse datasets featuring multiple scenes and corresponding anomalous conditions, like the UCF-Crime \cite{sultani2018real} and XD-Violence \cite{xdviolence} datasets.

Furthermore, these datasets exhibit variations in size and total duration, ranging from as low as five minutes to more than two hundred hours. Some datasets comprise a small number of long videos, primarily collected from fixed scenes, as seen in the Subway dataset \cite{Subway}, while others contain a larger number of shorter videos sourced from a wide array of locations, like the XD-Violence dataset \cite{xdviolence}. Additionally, the datasets showcase a noticeable diversity in the types of anomalies present, spanning from action-related anomalies such as unexpected movement direction and criminal activities to object-related anomalies like suspicious personnel and specific vehicle types.

Table \ref{tab:datasets} provides a comprehensive summary of the most commonly used publicly available datasets, while Figure \ref{fig:SampleFrames} illustrates sample frames extracted from these datasets.

Despite the availability of numerous datasets dedicated to the problem of VAD, several crucial trends and challenges need to be addressed:
\begin{itemize}
    \item The majority of the publicly available VAD datasets offer limited environmental diversity and are restricted to a specific setting. Examples are the ShanghaiTech \cite{ShanghaiTech} and the CUHK Avenue \cite{lu2013abnormal} datasets which are only limited to university campus scenes and anomalies. Such a trend can significantly hinder the trained models' capability to generalize well in other types of settings.

    \item {  A recurring theme in multiple datasets is the limited number of anomalous events contained in the dataset. Some datasets contained as little as three anomalous event types (strange action, wrong direction, abnormal object, CUHK Avenue \cite{lu2013abnormal}) in comparison to other datasets that featured up to 11 types of anomalies (ShanghaiTech \cite{ShanghaiTech}). This constraint lessens the models' efficacy in practical applications by limiting their capacity to learn a broad range of potential abnormal behaviors.}

    \item Most datasets available in the literature are imbalanced in terms of the normal and abnormal classes. Particularly in larger datasets like UCF-Crime \cite{sultani2018real}, the imbalance between normal and abnormal classes makes it difficult for models to be trained to detect anomalies accurately without being surpassed by the majority class (normal).
\end{itemize}

We draw the following future directions for the VAD problem in surveillance videos particularly in the realm of benchmark datasets for addressing the aforementioned issues: 
\begin{itemize}
    \item The benchmark datasets should be diverse and strive to encompass a wide range of anomalies, including both subtle and overt deviations from normal behavior. 
    \item The datasets should aim for realism, mimicking real-world surveillance environments as closely as possible. Additionally, scalability is crucial to accommodate the growing size and complexity of surveillance video data.
    \item  Anomalies often manifest over time, making temporal context essential for effective detection. 
Benchmark datasets should incorporate long-term temporal information to capture the dynamics of normal and abnormal behaviors accurately
\item Beyond detecting anomalies, future benchmark datasets could focus on localizing and segmenting anomalous regions within video frames or sequences. This finer granularity aids in understanding the nature and extent of anomalies, facilitating prompt response measures.
\item Integrating data from multiple modalities, such as visual, auditory, and textual information, can enhance anomaly detection performance. 
Future benchmark datasets might include multi-modal data to reflect the complexity of real-world surveillance systems.
\item For advancing anomaly detection research, openness and collaboration are crucial therefore future benchmark datasets should be openly accessible, inviting contributions from researchers worldwide and fostering innovation in the field.
\end{itemize}

By addressing these aspects, datasets can facilitate the development of more robust and effective anomaly detection approaches, ultimately improving the security and efficiency of surveillance systems.


\subsection{Feature Learning in Deep Video Anomaly Detection} \label{feature extraction}

Feature learning plays a pivotal role in the effectiveness of deep learning models for video anomaly detection. This section examines various feature extraction techniques utilized in the literature, emphasizing their significance and performance in surveillance video analysis.

\subsubsection{Exploring Different Feature Types}

Video frames encompass diverse types of features crucial for VAD. These include spatial, temporal, spatiotemporal, and textual features.
\paragraph{\textbf{Spatial Features}}

Spatial features pertain to the visual characteristics present within individual frames of a video. These encompass attributes such as shapes, textures, colors, and object positions within the frame.
In the realm of anomaly detection, the analysis of spatial features aids in the detection of unusual patterns or objects within specific regions of the video frame.
Initially, traditional machine learning methods dominated this area, employing techniques like Gaussian mixture models and manually constructed features \cite{HOF,HOG}. However, the transition to deep learning facilitated automated feature extraction, thereby augmenting the capability to discern intricate spatial details within video data.

\paragraph{\textbf{Temporal Features}}

Temporal features encompass changes or movements occurring over time within videos, including object motion, speed variations, and environmental alterations from frame to frame. For instance, Liu \textit{et al.} \cite{liu2018future} employed the optical flow approach to capture motion features within frames.
In video anomaly detection, temporal features play a pivotal role in identifying unusual actions or events spanning consecutive frames, such as unauthorized running in restricted areas \cite{sultani2018real,hasan2016learning}.

\paragraph{\textbf{Spatiotemporal Features}}

Relying solely on one type of feature can be restrictive: temporal features might fail to pinpoint where an anomaly occurs, while spatial features might be overlooked when it happens. Incorporating both spatiotemporal features offers a more holistic perspective, capturing not only the occurrence but also the precise location of an anomaly. This approach leads to more accurate and effective anomaly detection \cite{liu2018future,luo2021future,zhong2022bidirectional}.

\paragraph{\textbf{Textual Features}}

In video anomaly detection, textual features such as captioning and labeling significantly enhance the system's ability to recognize and understand anomalies. By incorporating descriptive captions and relevant labels into video frames, these systems gain a deeper understanding of the content, context, and actions depicted in the videos. This semantic layer aids in distinguishing normal from abnormal activities more effectively. Advanced techniques like vision language pretraining, including image-to-text, are employed to analyze these textual annotations, which can also encompass temporal and spatial context. The integration of textual features with visual data leads to more sophisticated, context-aware anomaly detection \cite{ni2022expanding,ju2022prompting,TEVAD,wu2023vadclip}.

\subsubsection{Deep Feature Extractors}
Different feature extractors have been utilized by previous researchers.

\paragraph{\textbf{Convolutional Neural Networks (CNNs)}}

\begin{itemize}
    \item  {2D Convolutional Neural Networks (2D CNNs):} CNNs have revolutionized spatial feature processing, enabling detailed analysis of structural elements in scenes. In their work \cite{mansour2021intelligent},  they discuss the use of Faster RCNN, a specific type of CNN architecture, because of its high accuracy and its dual capability of performing object classification and bounding box regression simultaneously. This means it can both locate and classify objects within the video frames, which is particularly useful for identifying and localizing anomalies.

\item {3D Convolutional Networks (3D CNNs): }  enhance traditional CNNs by incorporating temporal analysis, allowing for an effective evaluation of spatiotemporal characteristics within video data. The adoption of models such as C3D \cite{carreira2017quo} and I3D \cite{krizhevsky2012imagenet} have markedly elevated the performance in state-of-the-art (SOTA) systems. A significant number of studies, including those by \cite{sultani2018real,feng2021mist,zhong2019graph}, have employed these 3D CNN architectures as their foundational backbone, demonstrating their outstanding efficacy in spatiotemporal feature extraction.

\end{itemize}

\paragraph{\textbf{Autoencoders (AEs)}}

AEs excel in VAD due to their unsupervised learning ability. They encode data into a lower-dimensional space and then reconstruct it, learning key data features without needing labeled examples. This is vital in anomaly detection, where anomalies are rare and often not well-defined, allowing AEs to effectively identify unusual patterns in video data.

In the work of \cite{hasan2016learning}, their approach utilized a fully convolutional autoencoder to learn low-level motion features, enhancing the ability to learn regular dynamics in long-duration videos, making it an effective model for detecting irregularities. This method is versatile and applicable to various tasks, including temporal regularity analysis,  frame prediction, and abnormal event detection in videos.

\paragraph{\textbf{Generative Adversarial Networks (GANs)}}

GANs consist of two parts: a generator that creates realistic data and a discriminator that distinguishes between generated and real data. Through this adversarial process, GANs effectively learn the distribution of real data. This ability is particularly valuable in anomaly detection, as GANs can generate data that closely resembles normal instances, making it easier to identify anomalies that deviate from this learned pattern \cite{li2023multi}.

Their use in reconstruction-based approaches is especially noteworthy; these approaches reconstruct input data (like video snippets) using high-level representations learned from normal videos \cite{chen2022supervised}. The premise is that anomalies, being out-of-distribution inputs, are harder to reconstruct accurately compared to normal data, making reconstruction error a viable metric for anomaly detection.

As seen in the work of \cite{luo2021future}, GANs are employed for future frame predictions, which are then reconstructed, showcasing their versatility and effectiveness in anomaly detection tasks.

In this domain, GANs alongside AEs have shown to be effective in capturing these intricate patterns in video data, aiding in the more precise identification of anomalous activities. Their combined use enables the learning of high-level representations and the generation of realistic data, enhancing the ability to detect anomalies accurately and efficiently.

\paragraph{\textbf{Sequential Deep Learning}}
\begin{itemize}
 
 \item\textbf{Long Short-Term Memory (LSTM)}:
LSTMs are a specialized type of Recurrent Neural Network designed to capture temporal dependencies. This characteristic makes them exceptionally well-suited for processing sequential data, such as textual content in NLP-based systems \cite{abdalla2024nlp}. Moreover, their ability to remember and integrate long-range patterns is also invaluable in video applications. Specifically, they facilitate tracking and identifying temporal anomalies within sequential video streams, a critical function in spatiotemporal analysis \cite{nawaratne2019spatiotemporal}.

 \item \textbf{Vision Transformers (ViT)}: Vision Transformers, known for their attention mechanisms, have revolutionized the way sequential data is processed. They can weigh the importance of different parts of the input data, giving precedence to the most relevant features. This makes them highly effective for extracting both temporal and spatiotemporal features in videos, which is crucial for detecting complex anomalies. Yang \textit{et al.}\cite{yang2023video} introduce a novel video anomaly detection method by restoring events from keyframes, using a U-shaped Swin Transformer Network (a special type of ViT). This approach, distinct from traditional methods, focuses on inferring missing frames and capturing complex spatiotemporal relationships, demonstrating improved anomaly detection in video surveillance.

\end{itemize}

\paragraph{\textbf{Vision Language Models (VLM):}} Conventional techniques often rely solely on spatial-temporal characteristics, which might fall short in complex real-life situations where a deeper semantic insight is required for greater precision. Vision Language features involve training models using both visual and textual encoders, enabling comprehensive analysis of complex surveillance scenarios.

The notable rise of vision-language feature extractors using contrastive learning like the CLIP \cite{CLIP} and BLIP \cite{li2023blip} also aims to align vision and language, promising to bring about a transformative change in how surveillance videos are processed and interpreted. These models are designed to augment video content with rich semantic understanding, effectively narrowing the gap between simple pixel-based data and a more human-like interpretation of video content \cite{wu2023vadclip}, additionally, the language models using text captions are used in text-to-video anomaly retrieval as in \cite{10471334}. In their methodology of \cite{wu2023vadclip} VadClip model for video anomaly detection, Wu \textit{et al.} combined the CLIP model's vision-language features with a dual-branch system using both visual and textual encoders. It includes a local-global temporal adapter (LGT-Adapter) for modeling video temporal relations and leverages both visual and textual features for anomaly detection. One branch performs binary classification using visual features, while the other aligns visual and textual data for finer anomaly detection.

In contrast, \cite{TEVAD} produced text features using video captions with SwinBERT \cite{lin2022swinbert}, enriching the semantic context for detecting abnormalities. This approach expands the semantic understanding of video content beyond the pixel level, enhancing anomaly detection capabilities.

\paragraph{\textbf{Hybrid Feature Learners:}} Certain studies have developed hybrid feature extractors that integrate various types of feature extraction techniques to capture both spatial and temporal information effectively for video anomaly detection.

For example, \cite{yuan2021transanomaly} proposed a hybrid architecture that combines U-Net and a modified Video Vision Transformer (ViViT). The U-Net, known for its encoder-decoder structure with skip connections, excels at capturing detailed spatial information. In contrast, the modified ViViT, originally designed for video classification tasks, is adapted here to encode both spatial and temporal information effectively for video prediction. This hybrid model leverages the detailed high-resolution spatial information from the CNN features of the U-Net and the global context captured by the transformer.

Similarly, the work by \cite{transcnn} employed a composite system that combines CNNs with transformers. In this configuration, the CNN component is responsible for discerning spatial characteristics, whereas the transformers are tasked with identifying extended temporal attributes. This hybrid approach offers a complementary fusion of spatial and temporal features, enhancing the model's capability to detect anomalies in video sequences. Additionally, \cite{nguyen2019anomaly} proposed a CNN architecture that integrates convolutional autoencoders (CAEs) and a UNet. Each stream within the network contributes uniquely to the task of detecting anomalous frames. The CAEs are responsible for extracting spatial features, while the UNet focuses on capturing contextual information, enabling the model to effectively identify anomalies in video frames by leveraging both local and global features.

Similarly, \cite{majhi2020temporal} combined I3D 3D CNNs with LSTM networks to enhance anomaly detection performance. The I3D CNN extracts spatio-temporal features from video frames, effectively identifying potential anomalies. However, as CNNs primarily capture short-term dynamics, LSTMs are integrated to capture long-range temporal dependencies crucial in lengthy, untrimmed surveillance footage. To handle the high-dimensional data from the I3D network, a pooling strategy is employed to optimize feature vectors for efficient processing by the LSTM. This hybrid architecture effectively integrates the strengths of both CNNs and LSTMs, enhancing the model's ability to detect anomalies in varying temporal scales.

While each feature representation has its strengths and weaknesses as empirically investigated by the SOTA VAD approaches 
\cite{transcnn,CLIPTSA,wu2023vadclip,yang2023video,yang2024text}.
The end-to-end deep feature learning paradigms are the recent trends and these approaches have advanced the VAD performance.
We provide the following future directions in terms of the feature representations to improve detection, accuracy, and efficiency.

\begin{itemize}
\item A continuous exploration and refinement of deep learning architectures tailored for VAD is needed. 
This may involve developing more efficient architectures such as spatiotemporal CNNs, ViTs, or RNNs that can effectively capture temporal dependencies and spatial information in video sequences.
\item The hybrid models, which integrate multiple types of features, including appearance, motion, and spatiotemporal information, could also improve the performance.
This could involve combining deep learning-based features with handcrafted features or leveraging multi-modal approaches such as VLMs to capture diverse aspects of anomalies.
\item  Investigation of cross-modal feature representations that integrate information from multiple modalities, such as visual, audio, and textual cues present in surveillance videos. Cross-modal representations can capture complementary information sources and improve the robustness of anomaly detection models to diverse types of anomalies and environmental conditions.
\item Exploration of self-supervised learning techniques for pre-training feature representations in an unsupervised manner can help in learning rich representations from unlabeled data. The pre-trained models can then be fine-tuned for specific anomaly detection tasks with limited labeled data.
\item Integration of attention mechanisms into deep learning architectures to focus on relevant spatiotemporal regions or frames within videos. 
Attention mechanisms can help improve the model's ability to attend to informative parts of the input data and suppress irrelevant noise, leading to more robust anomaly detection features.
\item Utilization of graph-based feature representations to model complex relationships among different entities (e.g., objects, regions) within surveillance videos. GNNs can capture dependencies and interactions among entities, enabling more effective anomaly detection in scenarios where anomalies manifest as abnormal interactions or relationships.
\item Investigation of adversarial learning techniques to enhance the robustness of feature representations against adversarial attacks. 
Adversarial training methods can help improve the model's ability to generalize to unseen anomalies and mitigate the risk of evasion attacks in real-world surveillance systems.
\item Development of incremental learning approaches to adaptively update feature representations over time as new data becomes available can help the model adapt to evolving patterns of normal and abnormal behavior in dynamic surveillance environments without forgetting previously learned knowledge.
\end{itemize}


\subsection{Learning and Supervision Schemes} \label{DL methods}

In the context of this survey, we will delve into various learning approaches aimed at addressing the VAD problem.

   \subsubsection{Supervised Approaches}

In supervised learning, algorithms are developed using datasets where each frame is pre-labeled as either 'normal' or 'anomalous.' This allows the model to learn explicit distinctions between normal and abnormal events based on these annotations. However, the use of supervised methods in VAD is relatively rare. This is primarily because acquiring datasets with detailed, frame-level annotations is highly challenging and resource-intensive. Anomalous events are often rare and diverse, making it difficult to compile a comprehensive set of labeled anomalies. Additionally, the manual process of annotating each frame in extensive video data is laborious and time-consuming, further limiting the availability of such datasets. Consequently, while supervised learning can be powerful when precise labels are available, its application in VAD is constrained by the practical difficulties in obtaining extensive, accurately annotated training data.

\color{black}

An exemplary study by \cite{supervised2016} introduced an approach for detecting and localizing anomalies in crowded scenes using spatial-temporal Convolutional Neural Networks (CNNs). This method can simultaneously process spatial and temporal data from video sequences, capturing both appearance and motion information. Tailored specifically for crowded environments, it efficiently identifies anomalies by focusing on moving pixels, enhancing accuracy and robustness.

Another work by \cite{chen2022supervised} addresses the scarcity of labeled anomalies in anomaly detection by employing a conditional Generative Adversarial Network (cGAN) to generate supplementary training data that is class balanced. This approach utilizes labeled data for training the model. They propose a novel supervised anomaly detector, the Ensemble Active Learning Generative Adversarial Network (EAL-GAN). This network, unique in its architecture of one generator against multiple discriminators, leverages an ensemble learning loss function and an active learning algorithm. The goal is to alleviate the class imbalance problem and reduce the cost of labeling real-world data.

 \subsubsection{Self-supervised Approaches}
   Self-supervised approaches  
involve training models using data that is not explicitly labeled for anomalies. Instead, the models learn to identify anomalous events by solving "proxy tasks", which generate supervisory signals from the data itself. These tasks are designed to be related to the main goal of detecting anomalies, helping the model to develop an understanding of normal patterns in the data without needing direct supervision from labeled examples of anomalous events.  However, it demands careful design and selection of proxy tasks to ensure that the learned features are useful for identifying anomalous events.
 \color{black} 
In the work by \cite{georgescu2021anomaly}, researchers propose a method based on self-supervised and multi-task learning at the object level. Self-supervision involves training a 3D CNN on several proxy tasks that do not require labeled anomaly data. These tasks include determining the direction of object movement (arrow of time), identifying motion irregularities by comparing objects in consecutive versus intermittent frames, and reconstructing object appearances based on their preceding and succeeding frames. By learning from the video data itself what normal object behavior looks like, the model becomes adept at detecting anomalies when deviations from this learned behavior occur. This methodology enables effective anomaly detection even in the absence of explicit labels.

The authors continued their work in \cite{barbalau2023ssmtl++}. The new enhancements include integrating advanced object detection methods such as YOLOv5, optical flow, and background subtraction, which improve the detection of rapidly moving objects and those outside predefined classes. They also introduce transformer blocks into the architecture, exploring both 2D and 3D Convolutional Vision Transformers (CvT) to better capture complex spatiotemporal dependencies. These updates culminate in a more robust framework that significantly enhances the detection accuracy and adaptability in identifying anomalous events in video sequences.
\color{black}
\subsubsection{Weakly Supervised Approaches}

In weakly supervised approaches, unlike supervised approaches, obtaining precise frame-level annotations for anomalies in long video sequences can be challenging and time-consuming. Instead, annotators may label "snippets" or short segments within the video where anomalies are observed, serving as weak labels for training the model.

The concept of weak supervision was first introduced by Sultani \textit{et al.} \cite{sultani2018real} through their pioneering multiple instance learning (MIL) model. This model was the first to adopt the notion of weakly labeled training videos. In this approach, normal and anomalous videos are considered negative and positive bags, respectively, with video segments serving as instances in MIL. These bags are processed through feature extractors to capture spatiotemporal features before being directed to a fully connected network to produce the final output. The output anomaly score ranges from 0 to 1, with the goal of increasing the model's output anomaly score for abnormal segments while decreasing it for normal segments.
However, this method can introduce noisy labels since the weak annotations do not provide precise information about the exact location of anomalies within the segments. This ambiguity can lead to the model learning less accurate representations of normal and abnormal behaviors. Additionally, the reliance on segment-level labels can sometimes result in the inclusion of irrelevant frames within the labeled segments, further complicating the training process.

\color{black}

Other works have solved this issue. For example, the authors  \cite{zhong2019graph} introduced a novel approach to weakly supervised anomaly detection (WSVAD), reframing it as a supervised learning task with noisy labels, departing from the conventional MIL framework. Leveraging a Graph Convolutional Network (GCN), this method effectively cleans noisy labels, thereby enhancing the training and reliability of fully supervised action classifiers for anomaly detection. Additionally, the authors \cite{10268379}  introduced a binarization-embedded WSVAD (BE-WSVAD) method, which innovates by embedding binarization into GCN-based anomaly detection module.

In another study, \cite{zhang2019temporal} enhances traditional MIL with a Temporal Convolutional Network (TCN) and a unique Inner Bag Loss (IBL). The IBL strategically focuses on the variation of anomaly scores within each bag (video), emphasizing a larger score gap in positive bags (containing anomalies) and a smaller gap in negative bags (without anomalies). Meanwhile, the TCN effectively captures the temporal dynamics in videos, a crucial aspect often overlooked in standard MIL approaches.

Moreover, authors \cite{zaheer2020self,zaheer2021cleaning} propose a self-reasoning approach based on binary clustering of spatio-temporal video features to mitigate label noise in anomalous videos. Their framework involves generating pseudo labels through clustering, facilitating the cleaning of noise from the labels, and enhancing the overall anomaly detection performance. This method not only removes noisy labels but also improves the network's performance through a clustering distance loss.

Traditional MIL approaches often neglect the intricate interplay of features over time.

In \cite{purwanto2021dance}, the method commences with a relation-aware feature extractor that captures multi-scale CNN features from videos. The unique aspect of their approach is the integration of self-attention with Conditional Random Fields (CRFs), leveraging self-attention to capture short-range feature correlations and CRFs to learn feature interdependencies. This approach offers a more comprehensive analysis of complex movements and interactions for anomaly detection.

On the other hand, \cite{tian2021weakly} proposed Robust Temporal Feature Magnitude Learning (RTFM). RTFM addresses the challenge of recognizing rare abnormal snippets in videos dominated by normal events, particularly subtle anomalies that exhibit only minor differences from normal events. It employs temporal feature magnitude learning to improve the robustness of the MIL approach against negative instances from abnormal videos and integrates dilated convolutions and self-attention mechanisms to capture both long and short-range temporal dependencies.

Furthermore, \cite{feng2021mist} introduced ``MIST: Multiple Instance Self-Training Framework for Video Anomaly Detection" a novel approach for WSVAD. MIST diverges from traditional MIL by introducing a pseudo label generator with a sparse continuous sampling strategy for more accurate clip-level pseudo labels, and a self-guided attention-boosted feature encoder for focusing on anomalous regions within frames.

Additionally, \cite{zhang2022weakly} presents a novel weakly supervised temporal relation learning framework (WSTR). The framework, which uses I3D for feature extraction and incorporates snippet-level classifiers and top-k video classification for weak supervision, is the first of its kind to apply transformer technology in this context.

In \cite{CLIPTSA}, a novel approach called CLIPTSA leverages Vision Language (ViT) features for WSVAD. Unlike conventional models such as C3D or I3D, CLIPTSA utilizes the Vision Transformer (ViT) encoded visual features from CLIP \cite{CLIP} to efficiently extract discriminative representations. It incorporates a Temporal Self-Attention (TSA) mechanism to model both long- and short-range temporal dependencies, thereby enhancing detection performance in VAD.

Another recently published work by \cite{yang2024text}, proposes a novel weakly supervised framework called Text Prompt with Normality Guidance (TPWNG), leveraging the CLIP model for aligning textual descriptions with video frames to generate pseudo-labels. The method involves fine-tuning CLIP for domain adaptation using ranking and distributional inconsistency losses and introducing a learnable text prompt mechanism with normality visual prompts to improve text-video alignment. The framework also incorporates a pseudo-label generation module based on normality guidance to infer reliable frame-level pseudo-labels and a temporal context self-adaptive learning module to flexibly capture temporal dependencies in video events.
\color{black}
   \subsubsection{Unsupervised and Reconstruction-based Approaches}

Reconstruction-based approaches for video anomaly detection operate under the principle that normal events can be effectively reconstructed from a learned representation, while anomalies or abnormal events deviate significantly from this representation and hence are harder to reconstruct. Essentially, the model learns to represent or ``reconstruct" normal data, and anomalies are detected based on how poorly the model reconstructs them.

These methods are particularly suitable when labeled anomaly data is scarce. During training, only normal videos are considered, while during testing, the model is evaluated on both normal and abnormal videos to assess its anomaly detection capabilities.

Although these models are feasible, scalable, and cost-effective approaches. However, the effectiveness of these models is highly dependent on the quality and comprehensiveness of the normal training data. If the normal data is not representative of all possible normal variations, the model's performance in detecting anomalies can suffer.

Another issue is that the model could generate a high number of false positives, labeling normal variations or benign deviations as anomalies. This occurs because any deviation from the learned normal patterns, even if it is not truly anomalous, might be flagged.
\color{black}

 Deep learning techniques, especially CNN \cite{nguyen2019anomaly} or autoencoders \cite{hasan2016learning,chong2017abnormal}, are widely utilized for this approach. An autoencoder attempts to learn a compressed representation of its input data and then reconstructs the original data from this representation. During training, the model learns to minimize the reconstruction error between the input and the output.

After training, when the model encounters new data, it tries to reconstruct it based on what it has learned. The difference between the reconstructed output and the original input is measured, typically as a reconstruction error. A high reconstruction error indicates that the input is significantly different from what the model considers ``normal"  implying that the input might be an anomaly.

In the work by \cite{hasan2016learning}, the authors developed a method to assess the regularity of frames in video sequences using reconstruction models. They employed two autoencoders: one with convolutional layers and the other without. The model processed two distinct types of inputs: manually designed features (such as HOG and HOF, enhanced with trajectory-based characteristics), and a combination of 10 successive frames aligned along the time axis. The error in reconstructing these frames serves as an indicator of their regularity score.

\paragraph{The evolution of the reconstruction approaches}
The reconstruction-based paradigm is a key component of unsupervised learning frameworks and is often categorized within certain learning models as \textit{one-class classification (OCC)} or \textit{unsupervised learning}, primarily due to its emphasis on training using only the ``normal" class of videos. Within the OCC paradigm, the model is trained only on data from one class (typically the ``normal" class), and any deviation from this learned structure in the testing phase ``anomaly" results in higher reconstruction error.

On the other hand, unsupervised learning focuses on understanding the structure or distribution within the data itself. 
It learns to \textit{reconstruct} data without explicit labels indicating what is normal or anomalous.

As the work of, \cite{zaheer2022generative}, proposed a novel approach called Generative Cooperative Learning, which combines a generator and a discriminator trained together using a negative learning paradigm. The generator, designed as an autoencoder, reconstructs both normal and abnormal representations. Through negative learning, the discriminator learns to estimate the probability of an instance being abnormal by identifying the reconstruction errors from the generator. This method leverages the assumption that anomalies are less frequent than normal events and that normal events exhibit temporal consistency, allowing for more effective anomaly detection.

Another unsupervised approach by \cite{tur2023exploring} ) utilizes diffusion models, a type of generative model, to leverage the reconstruction capabilities for detecting anomalies. The approach begins by extracting features from video clips using a 3D convolutional neural network (3D-CNN). These features are then fed into a diffusion model, which reconstructs the features without relying on labeled data. The diffusion model progressively adds Gaussian noise to the input data and learns to reverse this process, effectively reconstructing the input.
\color{black}

  \textbf{Future Frame Prediction Approaches:} The concept of reconstruction-based approaches evolved as researchers observed that deep neural networks may not always produce significant reconstruction errors for unusual events. Additionally, certain normal videos, previously unencountered, could be incorrectly labeled as abnormal. To address these challenges, researchers began focusing on reconstructing future frames based on previous video frames while considering optical flow constraints to ensure motion consistency. This evolved methodology, termed ``prediction approaches," introduced Generative Adversarial Networks (GANs), which played a significant role in enhancing this approach \cite{liu2018future,luo2021future}. 
  Building upon this framework, the HSTGCNN \cite{9645572} model incorporates a sophisticated Future Frame Prediction (FFP) mechanism that significantly refines the anomaly detection process. By integrating hierarchical graph representations, the model not only predicts future frames but also encodes complex interactions among individuals and their movements, thus providing a more robust and context-aware anomaly detection system. 

A hybrid system incorporating flow reconstruction and frame prediction was introduced by \cite{liu2021hybrid}. This system detects unusual events in videos by memorizing normal activity patterns and predicting future frames. It utilizes a memory-augmented autoencoder for accurate flow reconstruction and a Conditional Variational Autoencoder for frame prediction. Anomalies are highlighted by larger errors in flow reconstruction and subsequent frame prediction.

Different VAD paradigms have their strengths and weaknesses. 
ViT-based approaches with the integration of language models/language supervision are the recent trends for improving VAD performance.
In the realm of different approaches for VAD, several promising directions are emerging:
\begin{itemize}
\item Development of vision-language models for VAD problems will be more dominant compared to vision-only models. These models do not only capture the semantic representation of the videos but also consider the natural language descriptions for anomalies.
\item Exploration of methods to localize and describe anomalies in videos using natural language descriptions is the recent trend. Language-guided anomaly localization techniques enable the model to identify spatial and temporal regions corresponding to anomalies and generate human-interpretable descriptions, enhancing situational awareness and facilitating response efforts.
    \item Exploration of GAN-based approaches for VAD, where the generator learns to generate normal video frames while the discriminator distinguishes between real and generated frames. GANs can capture complex distributions of normal behavior and detect deviations indicative of anomalies.
    \item Development of graph-based models that represent spatial and temporal dependencies among entities in surveillance videos as a graph structure. Spatiotemporal graph models can capture intricate relationships among objects, activities, and contexts, enabling more accurate anomaly detection in complex scenes.
    \item  Investigation of transfer learning techniques to transfer knowledge from labeled source domains to unlabeled or sparsely labeled target domains can help mitigate the scarcity of labeled data in target domains and improve anomaly detection performance by leveraging knowledge learned from related surveillance scenarios.
    \item  Adoption of multi-resolution analysis techniques to analyze videos at different spatial and temporal resolutions. Multi-resolution approaches enable the detection of anomalies occurring at varying scales, from small-scale events to large-scale spatial or temporal patterns, enhancing the model's sensitivity to diverse types of anomalies.
    \item Adoption of dynamic ensemble learning strategies to combine predictions from multiple anomaly detection models dynamically. 
    Dynamic ensemble methods can adaptively adjust the ensemble composition based on the current surveillance context, improving detection robustness and resilience to evolving anomalies.
\end{itemize}


\subsection{Loss Functions} \label{loss func}

Loss functions play a pivotal role in quantifying the disparity between predicted outcomes by a model and the actual results. They serve as guides for the optimization process, facilitating adjustments and enhancements to model parameters during training. The selection of an appropriate loss function is critical as it directly impacts the model's capacity to discern underlying patterns in the data and, consequently, its performance on unseen data. Various tasks necessitate distinct loss functions to adeptly capture the specific characteristics or complexities inherent in the problem domain.

\subsubsection{Multiple Instance Learning (MIL) Loss}

Within the category of weakly supervised learning, notably Multiple Instance Learning (MIL) \cite{sultani2018real}, the objective function is crafted to proficiently distinguish between normalcy and anomaly within video segments.

When dealing with a dataset of labeled videos,  denoted as  \(\mathcal{V}\), where  \(V\) is a video segment. The videos are categorized into positive bags \(\mathcal{B_+}\) and negative bags \(\mathcal{B_-}\). The overall objective function is represented as,
\vspace{-1.2\baselineskip}

\begin{equation}
\label{eq1}
\begin{aligned}
\argmin_{\mathcal{\theta}} & \sum_{\mathcal{B}_+, \mathcal{B}_- \in \mathcal{V}} \max \bigg(0, 1 - \max_{{V} \in \mathcal{B}_+} {S}({V};\mathcal{\theta}) + \max_{{V} \in \mathcal{B}_-} {S}({V};\mathcal{\theta}) \bigg)\\
\end{aligned}
\end{equation}
\vspace{-1\baselineskip}

This function aims to ensure that the anomaly score ${S}({V};\mathcal{\theta})$ for the positively labeled video dataset is greater than any segment within the negatively labeled set. This is achieved by employing a hinge loss function which is parameterized by \(\mathcal{\theta}\).

\subsubsection{Cross Entropy Loss}
Alongside the MIL loss, some other researchers   \cite{zhong2019graph, lv2023unbiased},   used the binary cross entropy within their loss functions to classify between normal and abnormal video snippets in the WSVAD tasks.  

The goal is to refine the model to identify the snippet with the highest anomaly score in a video, which is denoted as normal, and conversely in an anomalous video. For each video instance, MIL formulates a pair that includes the model's prediction for the snippet with the highest anomaly score and the true label for the video's anomaly  (e.g., \(\max \{f(V_i)\}_{i=1}^n, y_i\)), where \(y_i=0\) for normal and \(y_i=1\) for anomalous instances). MIL subsequently consolidates these pairs from all the videos to form a set  \(C\) of confidently labeled snippets. The model \(f\) is refined by optimizing the Binary Cross-Entropy (BCE) loss, calculated as follows:
\vspace{-1.2\baselineskip}

\begin{equation}
BCE(C) = -\frac{1}{N}\sum_{(y_i,\hat{y}_i) \in C} \left[ y_i \log(\hat{y}_i) + (1 - y_i) \log(1 - \hat{y}_i) \right]
\end{equation}

where the maximum score \(\hat{y}_i = \max \{ f(V_i) \}_{i=1}^n\). 
Under this paradigm, the model \(f\) must assign the lowest anomaly probability to all snippets in a normal video, thereby minimizing \(\max \{ f(V_i) \}_{i=1}^n\), and the highest anomaly probability in an anomalous video, hence maximizing \(\max \{ f(V_i) \}_{i=1}^n\). The strategy is to focus on the snippet with the strongest abnormality, even if the video is largely normal, to generate a set of snippets labeled with high confidence as anomalous.

\subsubsection{Reconstruction Error loss} As for the unsupervised approach, the objective function of the model, as shown in Equation \ref{eq2}, is to minimize the reconstruction error  (the squared Euclidean distance, \( \| \cdot \|_2^2 \)), as proposed by \cite{hasan2016learning}, between the original input frames pixel values and its reconstruction by the model with weights \( \mathcal{\theta} \).
The input frames denoted as \( V_i \) are passed through an encoder to obtain a compressed representation, which is then reconstructed back into frames \( F(V_i; \mathcal{\theta}) \) using a decoder and \(N\) is the size of the minibatch.


\begin{equation}
\label{eq2}
\begin{aligned}
\argmin_{\mathcal{\theta}} \frac{1}{2N} \sum_i \| V_i - {F}({V}_i;\mathcal{\theta}) \|_2^2 
\end{aligned}
\end{equation}

Future directions for video anomaly detection in terms of loss functions involve the exploration and development of novel loss functions tailored to address specific challenges and objectives in anomaly detection tasks. The promising future directions in this domain are:
\begin{itemize}
\item Designing anomaly-aware loss that explicitly accounts for the characteristics of anomalies could penalize model errors differently based on the severity or rarity of anomalies, helping the model focus more on detecting critical anomalies while reducing false alarms for common or benign events.
\item Incorporating temporal consistency constraints into the loss function to encourage smooth transitions and coherent predictions over time may penalize abrupt changes or inconsistencies in the model's predictions across consecutive frames, promoting more stable and coherent anomaly detection results.
\item Integrating adversarial losses into the training process to improve the robustness of anomaly detection models against adversarial attacks can encourage the model to generate predictions that are resilient to adversarial perturbations or manipulations, enhancing the reliability and effectiveness of anomaly detection systems in real-world scenarios.
\item Incorporating uncertainty-aware losses to quantify and mitigate uncertainty in anomaly detection predictions may enable the model to estimate the confidence or uncertainty associated with its predictions, facilitating more reliable decision-making and uncertainty quantification in anomaly detection systems.
\end{itemize}

\subsection{Regularization} \label{reg}

Regularization techniques in deep learning, particularly for video anomaly detection, are crucial for preventing overfitting and improving the generalization of the models. 

\subsubsection{Weight Decay regularization}

Weight decay is a form of regularization commonly used in training neural networks to prevent overfitting. The term ``weight decay" specifically refers to a technique that modifies the learning process to shrink the weights of the neural network during training. Weight decay works on the principle of adding a penalty to the loss function of a neural network related to the size of the weights \cite{kukavcka2017regularization}. The primary objective is to keep the weights small, which helps in reducing the model's complexity and its tendency to overfit the training data. By penalizing large weights, weight decay ensures that the model does not rely too heavily on any single feature or combination of features, leading to better generalization.

\paragraph{\textbf{L1 Regularization}}
L1 regularization is a regularization term that adds a penalty to the loss function of the model equivalent to the absolute value of the magnitude of the weights of the model. The general equation for a loss function with L1 regularization can be expressed generally as follows:

\begin{equation}
\label{eq3}
\begin{aligned}
L(\theta) = L_0(\theta) + \lambda \sum_{i=1}^{n} |\theta_i|
\end{aligned}
\end{equation}

where \( L(\theta) \) is the total loss function after including the L1 regularization term, \( L_0(\theta) \) is the original loss function without regularization, and \( \lambda \) is a regularization parameter that controls the strength of the regularization effect. Higher values of \( \lambda \) lead to greater regularization, potentially resulting in more features being effectively ignored by reducing their coefficients to zero. The key aspect of L1 regularization is its ability to create sparse models. Sparsity here refers to the property where some of the coefficients become exactly zero, effectively excluding some features from the model. This is particularly useful in feature selection, where the most important features in the model are to be identified. \\

\paragraph{\textbf{L2 Regularization}}
L2 regularization, also known as Ridge regularization, is a technique that modifies the loss function of a model by adding a penalty equivalent to the square of the magnitude of the model's weights. The formula for a loss function with L2 regularization is generally expressed as follows:

\begin{equation}
\label{eq4}
\begin{aligned}
L(\theta) = L_0(\theta) + \lambda \sum_{i=1}^{n} \theta_i^2
\end{aligned}
\end{equation}

Here, \( L(\theta) \) represents the total loss function, including the L2 regularization term. \( L_0(\theta) \) is the original loss function without regularization, and \( \lambda \) is the regularization parameter, which determines the strength of the regularization effect. Increasing the value of \( \lambda \) enhances the regularization effect, penalizing larger weights more significantly. The key characteristic of L2 regularization is its ability to prevent overfitting by keeping the weights small, which leads to a more generalized model. Unlike L1 regularization, which promotes sparsity in the model (some coefficients becoming exactly zero), L2 regularization tends to shrink all coefficients towards zero but does not make them exactly zero. This attribute makes it particularly useful in scenarios where many features contribute small amounts to the predictions. By penalizing the square of the weights, L2 regularization ensures that no single feature dominates the model, which can be essential for models where all features carry some level of importance. \\

\subsubsection{Temporal and Spatial Constraints}
Temporal and spatial constraints are other types of regularization terms that are commonly used in VAD \cite{sultani2018real}. The addition of these terms in the loss function ensures important specifications in the model's spatial and temporal learning for distinguishing between the normal and abnormal events in videos \cite{cho2023look}. Temporal constraints are used to guarantee the continuity and progression of events over time while spatial constraints are used to enhance the model's understanding of the objects' positions and physical locations throughout the frames of the video and their relation to the existence of anomalies. By incorporating both types of constraints in the loss function of a VAD model, the model is inclined toward achieving more robust anomaly detection.\\

\paragraph{\textbf{Temporal Smoothing Constraint}}

The temporal smoothing constraint is a crucial component of loss functions in VAD, employed to maintain stability in predicted anomaly scores across consecutive frames. This constraint aims to mitigate abrupt fluctuations in anomaly scores between adjacent frames, aligning with the expectation that successive frames typically exhibit similar anomaly characteristics.  The temporal smoothing constraint penalizes sudden changes in the anomaly score between sequential frames as follows in equation \ref{eq5}:

\begin{equation}
\label{eq5}
\begin{aligned}
L_{smooth} = \lambda_1 \sum_{i=1}^{|\mathcal{B}_+|} \bigg({S}({V}_i;\mathcal{\theta}) - {S}({V}_{i+1};\mathcal{\theta})\bigg)^2 
\end{aligned}
\end{equation}

where a $\lambda_1$ is a smoothing coefficient, ${S}({V}_i;\mathcal{\theta})$ is the anomaly score of frame $i$, and ${S}({V}_{i+1};\mathcal{\theta})$ is the anomaly score of frame $i+1$.\\

\paragraph{\textbf{Sparsity constraint}}

The sparsity constraint is a commonly used regularization term in VAD that forces the anomalous frames to be a minority among the video frames. In VAD, the anomalous frames are generally assumed to be the minority compared to the normal frames. The sparsity constraint penalizes the total number of anomalous frames in the video as follows in equation \ref{eq6}:

\begin{equation}
\label{eq6}
\begin{aligned}
L_{sparsity} = \lambda_2 \sum_{{V} \in \mathcal{B}_+} {S}({V}_i;\mathcal{\theta})
\end{aligned}
\end{equation}

\noindent where a $\lambda_2$ is a sparsity coefficient and ${S}({V}_i;\mathcal{\theta})$ is the anomaly score of frame $i$.\\

While spatio-temporal constraints suit well for handling the abrupt anomalies within the video sequence, the recent trends toward incorporating VAD-specific constraints improve the overall performance. Future directions in this domain include the following:

\begin{itemize}
    \item Adversarial regularization techniques may help the model learn more resilient features by introducing adversarial perturbations during training, thereby enhancing its ability to detect anomalies in the presence of adversarial manipulations.
    \item  Incorporating temporal regularization techniques to enforce temporal consistency and coherence in anomaly detection predictions over time can encourage smooth transitions and consistent predictions across consecutive frames.
    \item Graph-based regularization techniques can impose structural constraints on the learned representations, encouraging the model to capture meaningful interactions and contextual information among objects, scenes, or events, leading to more accurate anomaly detection.
    \item  Integrating sparse regularization techniques to encourage sparsity in the learned representations, focusing the model's attention on salient features or regions within videos can promote the selection of informative features while suppressing irrelevant noise.
    \item Addressing catastrophic forgetting and model degradation over time by incorporating continual learning regularization techniques can enable the model to adapt and learn from new data while preserving previously learned knowledge, facilitating the adaptation in the surveillance environments.
\end{itemize}

\subsection{Anomaly Score} \label{anomaly score}

The anomaly score indicates the likelihood of a segment or frame in a video being abnormal, calculated based on how much it deviates from normal patterns. A high anomaly score suggests a high probability of an anomaly.

For the reconstruction approaches \cite{hasan2016learning, chong2017abnormal}, after training the model, its effectiveness is measured by inputting test data to see if it can accurately identify unusual activities with a minimal number of false alarm rates. Then, calculate the anomaly score \(S(V_i)\) for each frame \(V_i\)  by scaling between 0 and 1 using the reconstruction error of the frame \(e(V_i)\).

\begin{equation}
S(V_i) =  \frac{e(V_i) - \min_{V_i} e(V_i)}{\max_{V_i} e(V_i)}.
\end{equation}

\noindent Hence, we can define the regularity score to be:
\begin{equation}
S_r(V_i) = 1 - S(V_i).
\end{equation}

 However for future frame prediction methods, \cite{mathieu2015deep} has shown that Peak Signal-to-Noise Ratio (PSNR)   is a superior method for evaluating image quality \cite{liu2018future, luo2021future}, as indicated below:


\begin{equation}
\text{PSNR}(Y, \hat{Y}) = 10 \log_{10} \left( \frac{\max_{\hat{Y}}^2}{\frac{1}{N} \sum_{i=0}^{N} (Y_i - \hat{Y}_{i})^2} \right),
\end{equation}

The assumption is that we can make accurate predictions for normal events. Therefore, we can detect anomalies by measuring the difference between a predicted frame \(\hat{Y}\) and its corresponding ground truth  \(Y\). Where \(\max_{\hat{Y}}\) is the maximum possible value of the image intensities divided by the mean square error (MSE). In image reconstruction tasks, a higher PSNR would indicate a lower error and, thus, higher fidelity to the original image, which is the desired outcome.

After calculating the PSNR between the predicted \(\hat{Y}\) and its corresponding ground truth  \(Y\) for each frame \(V_i\) in a test video and normalizing these values,   each frame's regularity score is determined \(S_r(V_i)\) using the equation below.
\begin{equation}
S_r(V_i)= \frac{\text{PSNR}(Y_{V_i}, \hat{Y}_{V_i}) - \min_{V_i}\text{PSNR}(Y_{V_i}, \hat{Y}_{V_i})}{\max_{V_i} \text{PSNR}(Y_{V_i}, \hat{Y}_{V_i}) -  \min_{V_i} \text{PSNR}(Y_{V_i}, \hat{Y}_{V_i})}
\end{equation}

This score indicates the likelihood of a frame being normal or abnormal, with the potential to set a threshold point for distinguishing between the two.

\section{Model Evaluation}
\label{expermentation and evaluation}

\subsection{Datasets Guideline}

As discussed in section \ref{datasets}, several datasets were used in literature for training and evaluating VAD models. These datasets are well-suited for different types of supervision schemes. Most of the datasets, including UCF-crime \cite{sultani2018real}, ShanghaiTech \cite{ShanghaiTech}, XD-Violence \cite{xdviolence}, and CUHK Avenue \cite{lu2013abnormal}, provide video-level labels for the training set which could be utilized for training weakly-supervised VAD as in the works by Joo \textit{et al.} \cite{CLIPTSA} and Cho \textit{et al.}\cite{cho2023look}. These datasets could also be utilized for unsupervised VAD by utilizing the training set without the video-level labels as in the works of Shi at al. \cite{shi2023video} and Zaheer \textit{et al.} \cite{zaheer2022generative} and for self-supervised learning as in the work of Zhou \textit{et al.} \cite{supervised2016}.


\subsection{Evaluation Metrics}
\label{evaluation}

Most previous works mainly compared their results on different datasets using metrics such as the Area Under the ROC Curve (AUC)  of the frame level, Equal Error Rate (EER), and Average Precision (AP) ) of the frame level to measure performance. The main aim of these metrics is to assess the model's proficiency in differentiating between normal and abnormal videos.

\textbf{Area Under the ROC Curve (AUC):}
The AUC is a significant metric used to evaluate model performance. It represents the area under the Receiver Operating Characteristic (ROC) curve, which plots the True Positive Rate (TPR) against the False Positive Rate (FPR) across various threshold settings. The TPR is defined as:
\begin{equation}
\text{TPR} = \frac{\text{TP}}{\text{TP} + \text{FN}}
\end{equation}
and the FPR is given by:
\begin{equation}
\text{FPR} = \frac{\text{FP}}{\text{FP} + \text{TN}}
\end{equation}
A higher AUC value indicates better model performance, where an AUC of 1 signifies a perfect model and an AUC of 0.5 suggests no discriminative power between positive and negative classes.
{ 
Two types of AUC that are commonly used for evaluating VAD models are Micro-AUC and Macro-AUC. Micro-AUC considers all samples across all classes together to compute a single AUC, making it sensitive to the most frequent classes. Macro-AUC, on the other hand, calculates the AUC for each class independently and averages these values, providing a balanced view of performance across all classes. While Micro-AUC reflects overall detection capability, Macro-AUC ensures the model performs well across different types of anomalies or normal events, offering insights into potential biases towards certain classes.}


\textbf{Average Precision (AP):}
The AP metric computes the average value of precision for different recall values. It summarizes the precision-recall curve into a single value representing average performance across all threshold levels.

\textbf{Equal Error Rate (EER):}
The EER represents the point at which the FPR equals the False Negative Rate (FNR). Lower EER values indicate a greater degree of accuracy in the system \cite{supervised2016}.

In this survey, we use the AUC as the performance metric in our quantitative comparison of the SOTA models in Section \ref{bench}.

\section{Comparative analysis of SOTA models} \label{comparative analysis}

\subsection{Quantitative comparison}\label{bench}
There have been several works in literature that proposed different architectures of deep learning models for encountering the problem of VAD. Table \ref{tab:bench} highlights a benchmarking of recent anomaly detection works on publicly available datasets. The table offers a detailed look at the evolution and current state of anomaly detection techniques, primarily focusing on publicly available datasets including UCF-Crime, Shanghai-Tech, XD-Violence, and Avenue. The data spans a decade, from 2013 to 2023, providing a comprehensive view of the field's progression in the last ten years. An important trend that can be observed in the table is the shift in research interest from supervised VAD approaches to weakly supervised VAD learning methods. This change, particularly notable from 2018 onwards and greatly influenced by the work of Sultani \textit{et al.} \cite{sultani2018real}  who introduced the ``UCF-Crime" dataset and the Multi-Instance Learning (MIL) learning pipeline, suggests a growing preference in the industry for techniques that require less fully labeled data, which is often expensive or difficult to obtain, in comparison to weakly supervised approaches that rely on video-level labels. This shift reflects the practical challenges and evolving needs in anomaly detection applications particularly in applications where hundreds of hours of unlabelled or weakly labeled video footage is available.\\

\begin{table*}[]
\centering
\caption{Benchmarking of recent anomaly detection works on publicly available datasets.}
\label{tab:bench}
\resizebox{\textwidth}{!}{%
\begin{tabular}{|l|l|l|l|c|c|c|c|c|l|}
\hline
\multicolumn{1}{|c|}{\textbf{Supervision type}} & \multicolumn{1}{c|}{\textbf{Method}} & \multicolumn{1}{c|}{\textbf{Venue}} & \multicolumn{1}{c|}{\textbf{Year}} & \multicolumn{1}{c|}{\textbf{Feature extractor}} & \textbf{UCF-Crime AUC} & \textbf{STech AUC} & \textbf{XD-Violance AUC} & \textbf{Avenue AUC} & \multicolumn{1}{c|}{\textbf{Ref}} \\ \hline
self-supervised & \textbf{SS-MTL} & CVPR & 2021 & 3D CNN & // & // & // & 86.9 & \cite{georgescu2021anomaly} \\ \hline
\multirow{12}{*}{Unsupervised} & \textbf{Lu \textit{et al.}} & ICCV & 2013 & 3D cube gradient and PCA & 65.51 & 68.00 & // & // & \cite{lu2013abnormal} \\ \cline{2-10} 
 & \textbf{Hasan \textit{et al.}} & CVPR & 2016 & Convolutional autoencoder & 50.60 & 60.85 & // & 70.20 & \cite{hasan2016learning} \\ \cline{2-10} 
 & \textbf{AMC} & ICCV & 2019 & Conv-AE & // & // & // & 86.9 & \cite{nguyen2019anomaly} \\ \cline{2-10} 
 & \textbf{MemAE} & ICCV & 2019 & 2D CNN Encoder-Decoder & // & 71.2 & // & 83.3 & \cite{gong2019memorizing} \\ \cline{2-10} 
 & \textbf{MPED-RNN} & CVPR & 2019 & RNN & // & 73.4 & // & // & \cite{morais2019learning} \\ \cline{2-10} 
 & \textbf{CVAE} & ICCV & 2021 & ML-MemAE-SC, CVAE & // & 76.2 & // & \textbf{91.1} & \cite{liu2021hybrid} \\ \cline{2-10} 
 & \textbf{GCL} & CVPR & 2022 & ResNext & 71.04 & \textbf{78.93} & // & // & \cite{zaheer2022generative} \\ \cline{2-10} 
 & \textbf{USTN-DSC} & CVPR & 2023 & Transformer Encoder-Decoder & // & 73.8 & // & 89.9 & \cite{yang2023video} \\ \cline{2-10} 
 & \textbf{FPDM} & ICCV & 2023 & Diffusion Model & \textbf{74.7} & 78.6 & // & // & \cite{yan2023feature} \\ \cline{2-10} 
 & \textbf{SLMPT} & ICCV & 2023 & U-Net & // & 78.8 & // & 90.9 & \cite{shi2023video} \\ \cline{2-10} 
 & \textbf{EVAL} & CVPR & 2023 & 3D CNN & // & 76.63 & // & 86.02 & \cite{singh2023eval} \\ \hline
\multirow{19}{*}{Weakly Sup.} & \textbf{Sultani \textit{et al.}} & CVPR & 2018 & C3D & 75.41 & // & // & // & \cite{sultani2018real} \\ \cline{2-10} 
 & \textbf{TCN-IBL} & ICIP & 2019 & C3D & 78.66 & 82.50 & // & // & \cite{zhang2019temporal} \\ \cline{2-10} 
 & \textbf{GCN-TSN} & CVPR & 2019 & TSN - RGB & 82.12 & 84.44 & // & // & \cite{zhong2019graph} \\ \cline{2-10} 
 & \textbf{SRF} & IEEE SPL & 2020 & C3D & 79.54 & 84.16 & // & // & \cite{zaheer2020self} \\ \cline{2-10} 
 & \textbf{Noise Cleaner} & CVPRW & 2021 & C3D & 78.26 & 84.16 & // & // & \cite{zaheer2021cleaning} \\ \cline{2-10} 
 & \textbf{DAM} & AVSS & 2021 & I3D- RGB & 82.67 & 88.22 & // & // & \cite{DAM} \\ \cline{2-10} 
 & \textbf{SA-CRF} & ICCV & 2021 & Relation-aware TSN ResNet-50 & 85.00 & 96.85 & // & // & \cite{purwanto2021dance} \\ \cline{2-10} 
 & \textbf{RTFM} & ICCV & 2021 & I3D- RGB & 84.30 & 97.21 & 77.81 & // & \cite{tian2021weakly} \\ \cline{2-10} 
 & \textbf{MIST} & CVPR & 2021 & I3D- RGB & 82.30 & 94.84 & // & // & \cite{feng2021mist} \\ \cline{2-10} 
 & \textbf{MSLNet-CTE} & AAAI & 2022 & VideoSwin-RGB & 85.62 & 97.32 & 78.59 & // & \cite{li2022self} \\ \cline{2-10} 
 & \textbf{CUPL} & CVPR & 2023 & I3D+VGGish & 86.22 & // & // & // & \cite{zhang2023exploiting} \\ \cline{2-10} 
 & \textbf{HSC} & CVPR & 2023 & AE & // & 83.4 & // & \textbf{93.7} & \cite{sun2023hierarchical} \\ \cline{2-10} 
 & \textbf{UMIL} & CVPR & 2023 & X-CLIP-B/32 & 86.75 & // & // & // & \cite{lv2023unbiased} \\ \cline{2-10} 
 & \textbf{TeD-SPAD} & ICCV & 2023 & U-Net + I3D & 75.06 & // & // & // & \cite{fioresi2023ted} \\ \cline{2-10} 
 & \textbf{CLAV} & CVPR & 2023 & I3D-RGB & 86.1 & 97.6 & 81.3 & 89.8 & \cite{cho2023look} \\ \cline{2-10} 
 & \textbf{TEVAD} & CVPRW & 2023 & ResNet-50 I3D-RGB / SwinBERT & 84.90 & 98.10 & 79.80 & // & \cite{TEVAD} \\ \cline{2-10} 
 & \textbf{CLIP-TSA} & ICIP & 2023 & CLIP & 87.58 & \textbf{98.32} & 82.19 & // & \cite{CLIPTSA} \\ \cline{2-10} 
 & \textbf{TPWNG} & CVPR & 2024 & CLIP (ViT-B/16) & \textbf{87.79} & // & \textbf{83.68} & // & \cite{yang2024text} \\ \hline
\end{tabular}%
}
\end{table*}

A notable aspect to consider is the wide array of feature extractors utilized in these studies. Ranging from simpler methods like 3D cube gradient with Principal Component Analysis (PCA) to more sophisticated deep learning architectures such as Convolutional Auto Encoders (CAE), 3D Convolutional Networks (C3D), Temporal Segment Networks (TSN), Inflated 3D ConvNet (I3D), and CLIP, the diversity in approaches underscores the intricate nature of the VAD task, emphasizing the necessity for tailored methodologies to address diverse scenarios.

\begin{figure*}[t] 
\centering
\frame{\includegraphics[width=\linewidth]{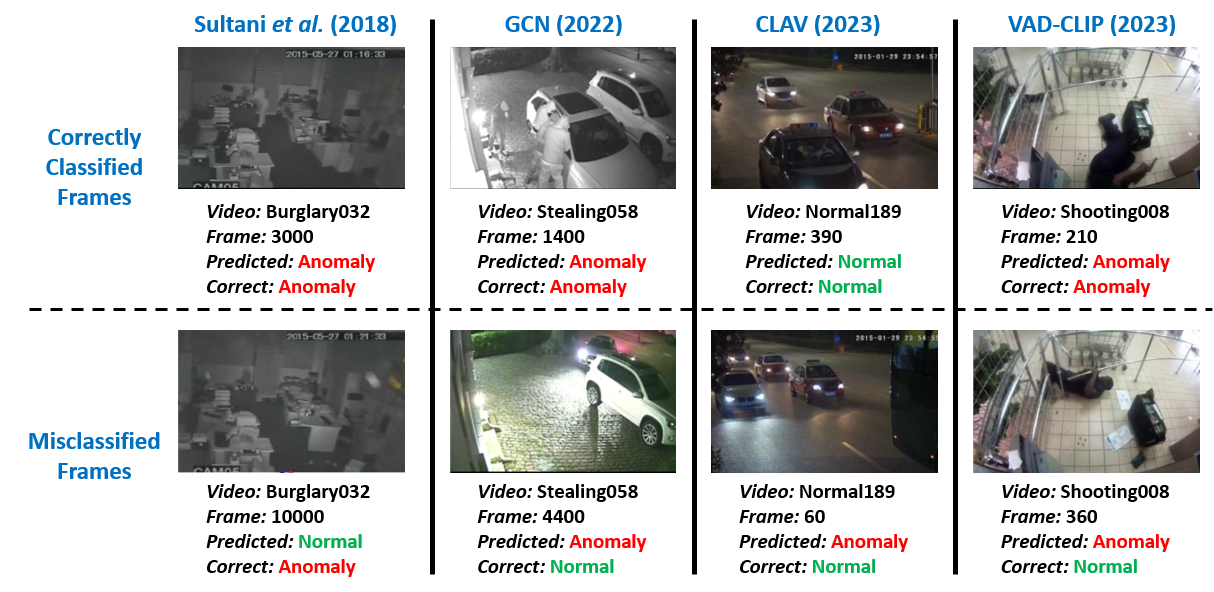}}

\caption{ \small 
 A qualitative comparison and illustration of correctly and incorrectly classified frames using four VAD models, namely Sultani \textit{et al.} \cite{sultani2018real}, GCN \cite{zaheer2022generative}, CLAV \cite{cho2023look}, and VAD-CLIP \cite{wu2023vadclip}. } 
\centering
\label{fig:Qualitative}
\end{figure*}

Another critical facet of the analysis pertains to performance metrics, particularly the Area Under the Curve (AUC) scores across different datasets, indicating notable enhancements in model performance over time. This trend not only signifies advancements in methodological efficacy but also underscores the increasing accuracy and reliability of anomaly detection systems. Leading models, notably recent ones like ``CLIP-TSA" \cite{CLIPTSA} published in 2023, exhibit exceptionally high AUC scores, illustrating the rapid progress achieved in recent years, particularly with the integration of advanced techniques from Natural Language Processing (NLP) into models like CLIP, BLIP, and GPT.
A novel approach that utilized the vision language models is \cite{yang2024text}, published in 2024, which outperformed the CLIP-TSA over two datasets as shown in Table \ref{tab:bench}. Based on the SOTA comparison, it is evident that Vision-language-based approaches are currently at the forefront of the VAD research.

\color{black}

However, the analysis reveals several instances of missing data. These gaps across various methodologies and datasets highlight the necessity for more comprehensive benchmarking efforts. Addressing such gaps is crucial to foster a deeper understanding and facilitate meaningful comparisons among different anomaly detection methods, ultimately advancing the field.


\subsection{Qualitative comparison} 

To observe in more detail the performance of current state-of-the-art models performance on various normal and anomalous surveillance videos, this section presents a qualitative analysis of these models' performances. Figure \ref{fig:Qualitative} depicts a qualitative assessment of correctly classified and misclassified video frames using four VAD models, namely Sultani \textit{et al.} \cite{sultani2018real}, GCN \cite{zaheer2022generative}, CLAV \cite{cho2023look}, and VAD-CLIP \cite{wu2023vadclip}. As shown in the figure's first row, these four VAD models were able to correctly classify normal and anomalous video frames taken from different scenes and environments. However, these models misclassified other similar frames taken from the same videos either by detecting anomalies in a normal video or by not detecting an anomalous video segment. For example, the model proposed by \cite{sultani2018real} misclassified the frame number 10000 in the video ``Burglary032", which shows a person entering an office through a window, by labeling it as ``Normal" while it is an ``Anomaly". This was attributed to the immense darkness of the scene which made recognizing such an ambiguous anomaly difficult. Another example of misclassification is with the GCN model by \cite{zaheer2022generative}  as it misclassified the frame number 4400 in the video ``Stealing058", which shows a car leaving the parking spot, by labeling it as an ``Anomaly" while it is  ``Normal". This could be attributed to the sudden movement of the car with its lights turned on. Similarly, the model proposed by \cite{cho2023look} named CLAV misclassified the initial frames of a scene showing cars stopping and accelerating again in a street. This increased anomaly score could be explained by the abrupt stopping of the cars which in many cases could be considered an anomaly. Finally, the VAD-CLIP model proposed by  \cite{wu2023vadclip} misclassified the frame number 360 in the video ``Shooting008", which shows a person crawling on the floor, by classifying it as an ``Anomaly" while it is ``Normal". This misclassification could be explained as the person crawling moves unconventionally in contrast with what commonly happens in normal street and shopping center scenes. It could be concluded that some normal and abnormal frames could be misclassified when the overall context of the video cannot be inferred directly from the frame in cases where the frames give a false impression of a different context.\\

\section{Visualizing Bibliometric Networks for Thematic Analysis} \label{Bibliometric}

In bibliometrics, visualization emerges as a potent technique for analyzing diverse networks, encompassing citation-based, co-authorship-based, or keyword co-occurrence-based networks. Density visualizations provide a rapid overview of the essential components within a bibliometric network. Employing NLP techniques, one can construct term co-occurrence networks for textual data in English. The rationale behind employing interconnected networks is to illustrate the temporal and relevance-driven evolution of the field.

We utilized VOSviewer \cite{van2010software}, a tool employing a distance-based method to generate visualizations of bibliometric networks. Directed networks, like those formed by citation relationships, are treated as undirected in this context. VOSviewer autonomously organizes network nodes into clusters, where nodes with similar attributes are grouped. Each node in the network is neatly assigned to one of these clusters. Furthermore, VOSviewer employs color to denote a node's membership within a cluster in bibliometric network visualizations.
\begin{figure*}[!h] 
\centering
\frame{\includegraphics[width=\linewidth]{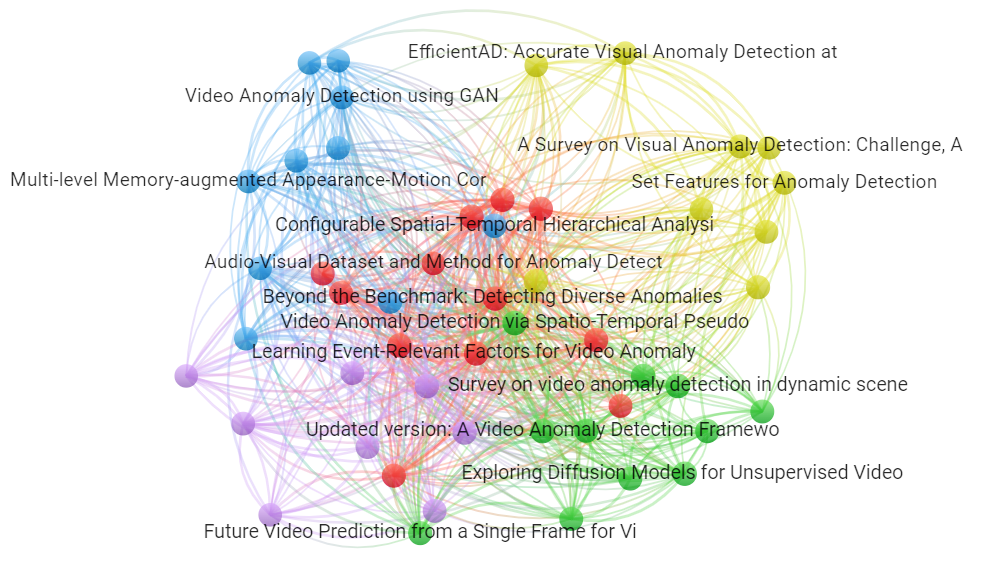}}

\caption{ Visualizing Bibliometric Networks for Thematic Analysis of Recent Literature (50 top cited papers) on Video Anomaly Detection between the year 2023-2024.
  }
\centering
\label{fig:Biometric}
\end{figure*}

Five distinct clusters have emerged, as shown in Figure \ref{fig:Biometric}, each representing a different research theme. The green cluster, centered around the theme of diffusion, explores methods based on diffusion models for anomaly detection, with representative work such as ``Anomaly Detection in Satellite Videos using Diffusion Models" \cite{awasthi2023anomaly}. In the yellow cluster, efficiency takes precedence, with research efforts directed towards developing algorithms capable of accurate anomaly detection at millisecond-level latencies, as demonstrated in the paper ``EfficientAD: Accurate Visual Anomaly Detection at Millisecond-Level Latencies" \cite{batzner2024efficientad}. The purple cluster focuses on interleaving one-class and weakly-supervised models with adaptive thresholding for unsupervised video anomaly detection, exemplified by the paper ``Interleaving One-Class and Weakly-Supervised Models with Adaptive Thresholding for Unsupervised Video Anomaly Detection" \cite{nie2024interleaving}. Within the red cluster, the exploration delves into vision language models (VLM), particularly CLIP, for video anomaly recognition, as seen in ``Delving into CLIP latent space for Video Anomaly Recognition" \cite{zanella2023delving}. Finally, the blue cluster is dedicated to research involving Generative Adversarial Networks (GANs) for VAD tasks, with representative work titled ``Video Anomaly Detection using GAN" \cite{sethi2023video}. All clusters are further illustrated in Table \ref{table3}.

\begin{table*}[ht]
\setlength\extrarowheight{2mm}
    \centering

\caption{Cluster central point and color for each cluster with representative paper and theme}
\label{table3}

\begin{tabularx}{\textwidth}{ | c | X |}
        \hline
         \textbf{{Cluster}} & \textbf{{Research Paper: The Title}} \\ \hline
 \statcirc{green} & Anomaly Detection in Satellite Videos using Diffusion Models (Theme: Diffusion) \\ \hline    
  \statcirc{red} & Delving into CLIP latent space for Video Anomaly Recognition (Theme: VLM) \\ \hline
 \statcirc{yellow} & EfficientAD: Accurate Visual Anomaly Detection at Millisecond-Level Latencies (Theme: Efficiency) \\ \hline
 \statcirc{blue} &  Video Anomaly Detection using GAN (Theme: GAN)\\ \hline
 \statcirc{mypurple} & Interleaving One-Class and Weakly-Supervised Models with Adaptive Thresholding for Unsupervised Video Anomaly Detection (Theme: Weakly-Supervised) \\ \hline
 \end{tabularx}
\end{table*}

\section{Discussion and Future Research} \label{conc}

In this survey paper, we present a comprehensive review that serves as a guideline to solve challenges related to VAD. The past decade has witnessed a remarkable evolution in the field of Video Anomaly Detection (VAD), marked by a notable transition from supervised to weakly supervised learning approaches and reconstruction-based techniques. This shift reflects a growing preference for methods capable of operating effectively with less reliance on fully labeled data, which can be both costly and challenging to obtain. Notably, the majority of benchmarking datasets are weakly supervised, featuring video-level annotations, while reconstruction approaches utilize only normal data in an unsupervised manner during training.

The existing literature presents a diverse array of VAD datasets, ranging from specific single-scene datasets like UCSD Pedestrian to more comprehensive, multi-scene collections such as UCF-Crime and XD-Violence. These datasets vary significantly in size, duration, and the nature of anomalies they encompass.

However, despite the progress made, challenges persist within the realm of VAD. Issues such as limited environmental diversity, a narrow range of anomalous event types, and class imbalance continue to impact the generalization and detection accuracy of models across different settings. This persistent gap underscores the urgent need for more diverse datasets, emphasizing the importance of expanding available resources to facilitate comprehensive and varied testing of VAD models.

Furthermore, the utilization of various hybrid deep learning techniques for feature extraction, including Convolutional Neural Networks (CNNs), Autoencoders (AEs), Generative Adversarial Networks (GANs), Sequential deep learning, and vision-language models as feature extractors, highlights the complexity of VAD tasks. These techniques underscore the extraction of crucial types of features such as spatiotemporal and textual features, underscoring the necessity for specialized approaches tailored to specific scenarios and reflecting the dynamic nature of the field.

Moreover, the selection of appropriate loss functions has played a pivotal role in the effectiveness of various tasks, serving as a fundamental component for model optimization and directly influencing a model's capacity to learn and make accurate predictions. Additionally, incorporating regularization terms into loss functions, particularly sparsity and smoothing constraints, has been essential to enhance the model's capability to discern between normal and abnormal events.

In the testing phase, evaluating the anomaly score using metrics such as Area Under the Curve (AUC), Average Precision (AP), and Equal Error Rate (EER) has been critical for comprehending the quality and biases of false positives, thereby providing insights into the model's limitations. It has been imperative to assess models across multiple datasets to ensure their robustness and generalization ability.

\subsection{Future Directions for Research}

Exploring the fusion of state-of-the-art vision-language models, particularly integrating textual features, with traditional VAD approaches presents a promising avenue for future investigations. This interdisciplinary approach holds the potential to enhance anomaly detection systems by imbuing them with a deeper understanding of complex video content, where semantic meanings are enriched through textual annotations.

While significant progress has been made in the field of VAD, there remains a pressing need for more diverse and extensive datasets. Specifically, datasets covering a broader range of scenarios, encompassing multiple scenes and anomalies, are essential. Such datasets would not only facilitate the benchmarking of existing models but also stimulate innovation by presenting researchers with more challenging real-world situations to address. 

Furthermore, since VLMs are emerging, an important direction for future research involves the integration of textual descriptions in the data containing contextual details, such as frame-level captions, into anomaly detection models. This integration has the potential to significantly improve model performance by providing rich, descriptive information that aids in the interpretation and analysis of visual content.
Consequently, this trend will necessitate the development of stronger loss functions capable of more effectively managing textual information.
In conclusion, this survey paper highlights an exciting phase of growth and transformation in the field of VAD, characterized by methodological advancements, the integration of new technologies, and a shift towards more efficient learning approaches. However, the identified gaps and challenges underscore the need for continued efforts within the research community to develop comprehensive datasets and explore novel methodologies, ultimately advancing the state of the art in anomaly detection.


\bibliographystyle{IEEEtran}

\bibliography{refs}

\begin{thebibliography}{10}
\providecommand{\url}[1]{#1}
\csname url@samestyle\endcsname
\providecommand{\newblock}{\relax}
\providecommand{\bibinfo}[2]{#2}
\providecommand{\BIBentrySTDinterwordspacing}{\spaceskip=0pt\relax}
\providecommand{\BIBentryALTinterwordstretchfactor}{4}
\providecommand{\BIBentryALTinterwordspacing}{\spaceskip=\fontdimen2\font plus
\BIBentryALTinterwordstretchfactor\fontdimen3\font minus \fontdimen4\font\relax}
\providecommand{\BIBforeignlanguage}[2]{{%
\expandafter\ifx\csname l@#1\endcsname\relax
\typeout{** WARNING: IEEEtran.bst: No hyphenation pattern has been}%
\typeout{** loaded for the language `#1'. Using the pattern for}%
\typeout{** the default language instead.}%
\else
\language=\csname l@#1\endcsname
\fi
#2}}
\providecommand{\BIBdecl}{\relax}
\BIBdecl

\bibitem{ramachandra2020survey}
B.~Ramachandra, M.~J. Jones, and R.~R. Vatsavai, ``A survey of single-scene video anomaly detection,'' \emph{IEEE transactions on pattern analysis and machine intelligence}, vol.~44, no.~5, pp. 2293--2312, 2020.

\bibitem{aggarwal2013applications}
C.~C. Aggarwal and C.~C. Aggarwal, ``Applications of outlier analysis,'' \emph{Outlier Analysis}, pp. 373--400, 2013.

\bibitem{jiang2023weakly}
M.~Jiang, C.~Hou, A.~Zheng, X.~Hu, S.~Han, H.~Huang, X.~He, P.~S. Yu, and Y.~Zhao, ``Weakly supervised anomaly detection: A survey,'' \emph{arXiv preprint arXiv:2302.04549}, 2023.

\bibitem{pang2021deep}
G.~Pang, C.~Shen, L.~Cao, and A.~V.~D. Hengel, ``Deep learning for anomaly detection: A review,'' \emph{ACM computing surveys (CSUR)}, vol.~54, no.~2, pp. 1--38, 2021.

\bibitem{shao2021transmil}
Z.~Shao, H.~Bian, Y.~Chen, Y.~Wang, J.~Zhang, X.~Ji \emph{et~al.}, ``Transmil: Transformer based correlated multiple instance learning for whole slide image classification,'' \emph{Advances in neural information processing systems}, vol.~34, pp. 2136--2147, 2021.

\bibitem{transcnn}
W.~Ullah, T.~Hussain, F.~U.~M. Ullah, M.~Y. Lee, and S.~W. Baik, ``Transcnn: Hybrid cnn and transformer mechanism for surveillance anomaly detection,'' \emph{Engineering Applications of Artificial Intelligence}, vol. 123, p. 106173, 2023.

\bibitem{nayak2021comprehensive}
R.~Nayak, U.~C. Pati, and S.~K. Das, ``A comprehensive review on deep learning-based methods for video anomaly detection,'' \emph{Image and Vision Computing}, vol. 106, p. 104078, 2021.

\bibitem{2018overview}
B.~R. Kiran, D.~M. Thomas, and R.~Parakkal, ``An overview of deep learning based methods for unsupervised and semi-supervised anomaly detection in videos,'' \emph{Journal of Imaging}, vol.~4, no.~2, p.~36, 2018.

\bibitem{georgescu2021anomaly}
M.-I. Georgescu, A.~Barbalau, R.~T. Ionescu, F.~S. Khan, M.~Popescu, and M.~Shah, ``Anomaly detection in video via self-supervised and multi-task learning,'' in \emph{Proceedings of the IEEE/CVF conference on computer vision and pattern recognition}, 2021, pp. 12\,742--12\,752.

\bibitem{liu2023unsupervised}
G.~Liu, L.~Shu, Y.~Yang, and C.~Jin, ``Unsupervised video anomaly detection in uavs: a new approach based on learning and inference,'' \emph{Frontiers in Sustainable Cities}, vol.~5, p. 1197434, 2023.

\bibitem{sultani2018real}
W.~Sultani, C.~Chen, and M.~Shah, ``Real-world anomaly detection in surveillance videos,'' in \emph{Proceedings of the IEEE conference on computer vision and pattern recognition}, 2018, pp. 6479--6488.

\bibitem{nawaratne2019spatiotemporal}
R.~Nawaratne, D.~Alahakoon, D.~De~Silva, and X.~Yu, ``Spatiotemporal anomaly detection using deep learning for real-time video surveillance,'' \emph{IEEE Transactions on Industrial Informatics}, vol.~16, no.~1, pp. 393--402, 2019.

\bibitem{xdviolence}
P.~Wu, J.~Liu, Y.~Shi, Y.~Sun, F.~Shao, Z.~Wu, and Z.~Yang, ``Not only look, but also listen: Learning multimodal violence detection under weak supervision,'' in \emph{Computer Vision--ECCV 2020: 16th European Conference, Glasgow, UK, August 23--28, 2020, Proceedings, Part XXX 16}.\hskip 1em plus 0.5em minus 0.4em\relax Springer, 2020, pp. 322--339.

\bibitem{ShanghaiTech}
W.~Luo, W.~Liu, and S.~Gao, ``A revisit of sparse coding based anomaly detection in stacked rnn framework,'' in \emph{2017 IEEE International Conference on Computer Vision (ICCV)}, 2017, pp. 341--349.

\bibitem{lu2013abnormal}
C.~Lu, J.~Shi, and J.~Jia, ``Abnormal event detection at 150 fps in matlab,'' in \emph{Proceedings of the IEEE international conference on computer vision}, 2013, pp. 2720--2727.

\bibitem{hasan2016learning}
M.~Hasan, J.~Choi, J.~Neumann, A.~K. Roy-Chowdhury, and L.~S. Davis, ``Learning temporal regularity in video sequences,'' in \emph{Proceedings of the IEEE conference on computer vision and pattern recognition}, 2016, pp. 733--742.

\bibitem{HOG}
N.~Dalal and B.~Triggs, ``Histograms of oriented gradients for human detection,'' in \emph{2005 IEEE computer society conference on computer vision and pattern recognition (CVPR'05)}, vol.~1.\hskip 1em plus 0.5em minus 0.4em\relax Ieee, 2005, pp. 886--893.

\bibitem{HOF}
N.~Dalal, B.~Triggs, and C.~Schmid, ``Human detection using oriented histograms of flow and appearance,'' in \emph{Computer Vision--ECCV 2006: 9th European Conference on Computer Vision, Graz, Austria, May 7-13, 2006. Proceedings, Part II 9}.\hskip 1em plus 0.5em minus 0.4em\relax Springer, 2006, pp. 428--441.

\bibitem{luo2017remembering}
W.~Luo, W.~Liu, and S.~Gao, ``Remembering history with convolutional lstm for anomaly detection,'' in \emph{2017 IEEE International conference on multimedia and expo (ICME)}.\hskip 1em plus 0.5em minus 0.4em\relax IEEE, 2017, pp. 439--444.

\bibitem{liu2021hybrid}
Z.~Liu, Y.~Nie, C.~Long, Q.~Zhang, and G.~Li, ``A hybrid video anomaly detection framework via memory-augmented flow reconstruction and flow-guided frame prediction,'' in \emph{Proceedings of the IEEE/CVF international conference on computer vision}, 2021, pp. 13\,588--13\,597.

\bibitem{li2023multi}
D.~Li, X.~Nie, R.~Gong, X.~Lin, and H.~Yu, ``Multi-branch gan-based abnormal events detection via context learning in surveillance videos,'' \emph{IEEE Transactions on Circuits and Systems for Video Technology}, 2023.

\bibitem{sun2023hierarchical}
S.~Sun and X.~Gong, ``Hierarchical semantic contrast for scene-aware video anomaly detection,'' in \emph{Proceedings of the IEEE/CVF Conference on Computer Vision and Pattern Recognition}, 2023, pp. 22\,846--22\,856.

\bibitem{li2022self}
S.~Li, F.~Liu, and L.~Jiao, ``Self-training multi-sequence learning with transformer for weakly supervised video anomaly detection,'' in \emph{Proceedings of the AAAI Conference on Artificial Intelligence}, vol.~36, no.~2, 2022, pp. 1395--1403.

\bibitem{ni2022expanding}
B.~Ni, H.~Peng, M.~Chen, S.~Zhang, G.~Meng, J.~Fu, S.~Xiang, and H.~Ling, ``Expanding language-image pretrained models for general video recognition,'' in \emph{European Conference on Computer Vision}.\hskip 1em plus 0.5em minus 0.4em\relax Springer, 2022, pp. 1--18.

\bibitem{ju2022prompting}
C.~Ju, T.~Han, K.~Zheng, Y.~Zhang, and W.~Xie, ``Prompting visual-language models for efficient video understanding,'' in \emph{European Conference on Computer Vision}.\hskip 1em plus 0.5em minus 0.4em\relax Springer, 2022, pp. 105--124.

\bibitem{CLIPTSA}
H.~K. Joo, K.~Vo, K.~Yamazaki, and N.~Le, ``Clip-tsa: Clip-assisted temporal self-attention for weakly-supervised video anomaly detection,'' in \emph{2023 IEEE International Conference on Image Processing (ICIP)}.\hskip 1em plus 0.5em minus 0.4em\relax IEEE, 2023, pp. 3230--3234.

\bibitem{zhang2019temporal}
J.~Zhang, L.~Qing, and J.~Miao, ``Temporal convolutional network with complementary inner bag loss for weakly supervised anomaly detection,'' in \emph{2019 IEEE International Conference on Image Processing (ICIP)}.\hskip 1em plus 0.5em minus 0.4em\relax IEEE, 2019, pp. 4030--4034.

\bibitem{zhong2019graph}
J.-X. Zhong, N.~Li, W.~Kong, S.~Liu, T.~H. Li, and G.~Li, ``Graph convolutional label noise cleaner: Train a plug-and-play action classifier for anomaly detection,'' in \emph{Proceedings of the IEEE/CVF conference on computer vision and pattern recognition}, 2019, pp. 1237--1246.

\bibitem{zaheer2020self}
M.~Z. Zaheer, A.~Mahmood, H.~Shin, and S.-I. Lee, ``A self-reasoning framework for anomaly detection using video-level labels,'' \emph{IEEE Signal Processing Letters}, vol.~27, pp. 1705--1709, 2020.

\bibitem{DAM}
S.~Majhi, S.~Das, and F.~Br{\'e}mond, ``Dam: dissimilarity attention module for weakly-supervised video anomaly detection,'' in \emph{2021 17th IEEE International Conference on Advanced Video and Signal Based Surveillance (AVSS)}.\hskip 1em plus 0.5em minus 0.4em\relax IEEE, 2021, pp. 1--8.

\bibitem{purwanto2021dance}
D.~Purwanto, Y.-T. Chen, and W.-H. Fang, ``Dance with self-attention: A new look of conditional random fields on anomaly detection in videos,'' in \emph{Proceedings of the IEEE/CVF International Conference on Computer Vision}, 2021, pp. 173--183.

\bibitem{tian2021weakly}
Y.~Tian, G.~Pang, Y.~Chen, R.~Singh, J.~W. Verjans, and G.~Carneiro, ``Weakly-supervised video anomaly detection with robust temporal feature magnitude learning,'' in \emph{Proceedings of the IEEE/CVF international conference on computer vision}, 2021, pp. 4975--4986.

\bibitem{TEVAD}
W.~Chen, K.~T. Ma, Z.~J. Yew, M.~Hur, and D.~A.-A. Khoo, ``Tevad: Improved video anomaly detection with captions,'' in \emph{Proceedings of the IEEE/CVF Conference on Computer Vision and Pattern Recognition}, 2023, pp. 5548--5558.

\bibitem{CLIP}
A.~Radford, J.~W. Kim, C.~Hallacy, A.~Ramesh, G.~Goh, S.~Agarwal, G.~Sastry, A.~Askell, P.~Mishkin, J.~Clark \emph{et~al.}, ``Learning transferable visual models from natural language supervision,'' in \emph{International conference on machine learning}.\hskip 1em plus 0.5em minus 0.4em\relax PMLR, 2021, pp. 8748--8763.

\bibitem{wu2023vadclip}
P.~Wu, X.~Zhou, G.~Pang, L.~Zhou, Q.~Yan, P.~Wang, and Y.~Zhang, ``Vadclip: Adapting vision-language models for weakly supervised video anomaly detection,'' in \emph{Proceedings of the AAAI Conference on Artificial Intelligence}, vol.~38, no.~6, 2024, pp. 6074--6082.

\bibitem{lin2022swinbert}
K.~Lin, L.~Li, C.-C. Lin, F.~Ahmed, Z.~Gan, Z.~Liu, Y.~Lu, and L.~Wang, ``Swinbert: End-to-end transformers with sparse attention for video captioning,'' in \emph{Proceedings of the IEEE/CVF Conference on Computer Vision and Pattern Recognition}, 2022, pp. 17\,949--17\,958.

\bibitem{chalapathy2019deep}
R.~Chalapathy and S.~Chawla, ``Deep learning for anomaly detection: A survey,'' \emph{arXiv preprint arXiv:1901.03407}, 2019.

\bibitem{baradaran2024critical}
M.~Baradaran and R.~Bergevin, ``A critical study on the recent deep learning based semi-supervised video anomaly detection methods,'' \emph{Multimedia Tools and Applications}, vol.~83, no.~9, pp. 27\,761--27\,807, 2024.

\bibitem{choudhry2023comprehensive}
N.~Choudhry, J.~Abawajy, S.~Huda, and I.~Rao, ``A comprehensive survey of machine learning methods for surveillance videos anomaly detection,'' \emph{IEEE Access}, 2023.

\bibitem{liu2023generalized}
Y.~Liu, D.~Yang, Y.~Wang, J.~Liu, J.~Liu, A.~Boukerche, P.~Sun, and L.~Song, ``Generalized video anomaly event detection: Systematic taxonomy and comparison of deep models,'' \emph{ACM Computing Surveys}, 2023.

\bibitem{Subway}
A.~Adam, E.~Rivlin, I.~Shimshoni, and D.~Reinitz, ``Robust real-time unusual event detection using multiple fixed-location monitors,'' \emph{IEEE Transactions on Pattern Analysis and Machine Intelligence}, vol.~30, no.~3, pp. 555--560, 2008.

\bibitem{UCSD}
V.~Mahadevan, W.~Li, V.~Bhalodia, and N.~Vasconcelos, ``Anomaly detection in crowded scenes,'' in \emph{2010 IEEE Computer Society Conference on Computer Vision and Pattern Recognition}, 2010, pp. 1975--1981.

\bibitem{ramachandra2020street}
B.~Ramachandra and M.~Jones, ``Street scene: A new dataset and evaluation protocol for video anomaly detection,'' in \emph{Proceedings of the IEEE/CVF Winter Conference on Applications of Computer Vision}, 2020, pp. 2569--2578.

\bibitem{singh2023eval}
A.~Singh, M.~J. Jones, and E.~G. Learned-Miller, ``Eval: Explainable video anomaly localization,'' in \emph{Proceedings of the IEEE/CVF Conference on Computer Vision and Pattern Recognition}, 2023, pp. 18\,717--18\,726.

\bibitem{NWPU}
C.~Cao, Y.~Lu, P.~Wang, and Y.~Zhang, ``A new comprehensive benchmark for semi-supervised video anomaly detection and anticipation,'' in \emph{Proceedings of the IEEE/CVF Conference on Computer Vision and Pattern Recognition (CVPR)}, June 2023, pp. 20\,392--20\,401.

\bibitem{liu2018future}
W.~Liu, W.~Luo, D.~Lian, and S.~Gao, ``Future frame prediction for anomaly detection–a new baseline,'' in \emph{Proceedings of the IEEE Conference on Computer Vision and Pattern Recognition (CVPR)}, 2018, pp. 6536--6545.

\bibitem{luo2021future}
W.~Luo, W.~Liu, D.~Lian, and S.~Gao, ``Future frame prediction network for video anomaly detection,'' \emph{IEEE transactions on pattern analysis and machine intelligence}, vol.~44, no.~11, pp. 7505--7520, 2021.

\bibitem{zhong2022bidirectional}
Y.~Zhong, X.~Chen, Y.~Hu, P.~Tang, and F.~Ren, ``Bidirectional spatio-temporal feature learning with multiscale evaluation for video anomaly detection,'' \emph{IEEE Transactions on Circuits and Systems for Video Technology}, vol.~32, no.~12, pp. 8285--8296, 2022.

\bibitem{mansour2021intelligent}
R.~F. Mansour, J.~Escorcia-Gutierrez, M.~Gamarra, J.~A. Villanueva, and N.~Leal, ``Intelligent video anomaly detection and classification using faster rcnn with deep reinforcement learning model,'' \emph{Image and Vision Computing}, vol. 112, p. 104229, 2021.

\bibitem{carreira2017quo}
J.~Carreira and A.~Zisserman, ``Quo vadis, action recognition? a new model and the kinetics dataset,'' in \emph{proceedings of the IEEE Conference on Computer Vision and Pattern Recognition}, 2017, pp. 6299--6308.

\bibitem{krizhevsky2012imagenet}
A.~Krizhevsky, I.~Sutskever, and G.~E. Hinton, ``Imagenet classification with deep convolutional neural networks,'' \emph{Advances in neural information processing systems}, vol.~25, 2012.

\bibitem{feng2021mist}
J.-C. Feng, F.-T. Hong, and W.-S. Zheng, ``Mist: Multiple instance self-training framework for video anomaly detection,'' in \emph{Proceedings of the IEEE/CVF conference on computer vision and pattern recognition}, 2021, pp. 14\,009--14\,018.

\bibitem{chen2022supervised}
Z.~Chen, J.~Duan, L.~Kang, and G.~Qiu, ``Supervised anomaly detection via conditional generative adversarial network and ensemble active learning,'' \emph{IEEE Transactions on Pattern Analysis and Machine Intelligence}, vol.~45, no.~6, pp. 7781--7798, 2022.

\bibitem{abdalla2024nlp}
M.~Abdalla, H.~Hassan, N.~Mostafa, S.~Abdelghafar, A.~Al-Kabbany, and M.~Hadhoud, ``An nlp-based system for modulating virtual experiences using speech instructions,'' \emph{Expert Systems with Applications}, vol. 249, p. 123484, 2024.

\bibitem{yang2023video}
Z.~Yang, J.~Liu, Z.~Wu, P.~Wu, and X.~Liu, ``Video event restoration based on keyframes for video anomaly detection,'' in \emph{Proceedings of the IEEE/CVF Conference on Computer Vision and Pattern Recognition}, 2023, pp. 14\,592--14\,601.

\bibitem{li2023blip}
J.~Li, D.~Li, S.~Savarese, and S.~Hoi, ``Blip-2: Bootstrapping language-image pre-training with frozen image encoders and large language models,'' \emph{arXiv preprint arXiv:2301.12597}, 2023.

\bibitem{10471334}
P.~Wu, J.~Liu, X.~He, Y.~Peng, P.~Wang, and Y.~Zhang, ``Toward video anomaly retrieval from video anomaly detection: New benchmarks and model,'' \emph{IEEE Transactions on Image Processing}, vol.~33, pp. 2213--2225, 2024.

\bibitem{yuan2021transanomaly}
H.~Yuan, Z.~Cai, H.~Zhou, Y.~Wang, and X.~Chen, ``Transanomaly: Video anomaly detection using video vision transformer,'' \emph{IEEE Access}, vol.~9, pp. 123\,977--123\,986, 2021.

\bibitem{nguyen2019anomaly}
T.-N. Nguyen and J.~Meunier, ``Anomaly detection in video sequence with appearance-motion correspondence,'' in \emph{Proceedings of the IEEE/CVF international conference on computer vision}, 2019, pp. 1273--1283.

\bibitem{majhi2020temporal}
S.~Majhi, R.~Dash, and P.~K. Sa, ``Temporal pooling in inflated 3dcnn for weakly-supervised video anomaly detection,'' in \emph{2020 11th International Conference on Computing, Communication and Networking Technologies (ICCCNT)}.\hskip 1em plus 0.5em minus 0.4em\relax IEEE, 2020, pp. 1--6.

\bibitem{yang2024text}
Z.~Yang, J.~Liu, and P.~Wu, ``Text prompt with normality guidance for weakly supervised video anomaly detection,'' in \emph{Proceedings of the IEEE/CVF Conference on Computer Vision and Pattern Recognition}, 2024, pp. 18\,899--18\,908.

\bibitem{supervised2016}
S.~Zhou, W.~Shen, D.~Zeng, M.~Fang, Y.~Wei, and Z.~Zhang, ``Spatial--temporal convolutional neural networks for anomaly detection and localization in crowded scenes,'' \emph{Signal Processing: Image Communication}, vol.~47, pp. 358--368, 2016.

\bibitem{barbalau2023ssmtl++}
A.~Barbalau, R.~T. Ionescu, M.-I. Georgescu, J.~Dueholm, B.~Ramachandra, K.~Nasrollahi, F.~S. Khan, T.~B. Moeslund, and M.~Shah, ``Ssmtl++: Revisiting self-supervised multi-task learning for video anomaly detection,'' \emph{Computer Vision and Image Understanding}, vol. 229, p. 103656, 2023.

\bibitem{10268379}
Z.~Yang, Y.~Guo, J.~Wang, D.~Huang, X.~Bao, and Y.~Wang, ``Towards video anomaly detection in the real world: A binarization embedded weakly-supervised network,'' \emph{IEEE Transactions on Circuits and Systems for Video Technology}, pp. 1--1, 2023.

\bibitem{zaheer2021cleaning}
M.~Z. Zaheer, J.-h. Lee, M.~Astrid, A.~Mahmood, and S.-I. Lee, ``Cleaning label noise with clusters for minimally supervised anomaly detection,'' \emph{arXiv preprint arXiv:2104.14770}, 2021.

\bibitem{zhang2022weakly}
D.~Zhang, C.~Huang, C.~Liu, and Y.~Xu, ``Weakly supervised video anomaly detection via transformer-enabled temporal relation learning,'' \emph{IEEE Signal Processing Letters}, vol.~29, pp. 1197--1201, 2022.

\bibitem{chong2017abnormal}
Y.~S. Chong and Y.~H. Tay, ``Abnormal event detection in videos using spatiotemporal autoencoder,'' in \emph{Advances in Neural Networks-ISNN 2017: 14th International Symposium, ISNN 2017, Sapporo, Hakodate, and Muroran, Hokkaido, Japan, June 21–26, 2017, Proceedings, Part II}.\hskip 1em plus 0.5em minus 0.4em\relax Springer International Publishing, 2017, pp. 189--196.

\bibitem{zaheer2022generative}
M.~Z. Zaheer, A.~Mahmood, M.~H. Khan, M.~Segu, F.~Yu, and S.-I. Lee, ``Generative cooperative learning for unsupervised video anomaly detection,'' in \emph{Proceedings of the IEEE/CVF conference on computer vision and pattern recognition}, 2022, pp. 14\,744--14\,754.

\bibitem{tur2023exploring}
A.~O. Tur, N.~Dall’Asen, C.~Beyan, and E.~Ricci, ``Exploring diffusion models for unsupervised video anomaly detection,'' in \emph{2023 IEEE International Conference on Image Processing (ICIP)}.\hskip 1em plus 0.5em minus 0.4em\relax IEEE, 2023, pp. 2540--2544.

\bibitem{9645572}
X.~Zeng, Y.~Jiang, W.~Ding, H.~Li, Y.~Hao, and Z.~Qiu, ``A hierarchical spatio-temporal graph convolutional neural network for anomaly detection in videos,'' \emph{IEEE Transactions on Circuits and Systems for Video Technology}, vol.~33, no.~1, pp. 200--212, 2023.

\bibitem{lv2023unbiased}
H.~Lv, Z.~Yue, Q.~Sun, B.~Luo, Z.~Cui, and H.~Zhang, ``Unbiased multiple instance learning for weakly supervised video anomaly detection,'' in \emph{Proceedings of the IEEE/CVF Conference on Computer Vision and Pattern Recognition}, 2023, pp. 8022--8031.

\bibitem{kukavcka2017regularization}
J.~Kuka{\v{c}}ka, V.~Golkov, and D.~Cremers, ``Regularization for deep learning: A taxonomy,'' \emph{arXiv preprint arXiv:1710.10686}, 2017.

\bibitem{cho2023look}
M.~Cho, M.~Kim, S.~Hwang, C.~Park, K.~Lee, and S.~Lee, ``Look around for anomalies: Weakly-supervised anomaly detection via context-motion relational learning,'' in \emph{Proceedings of the IEEE/CVF Conference on Computer Vision and Pattern Recognition}, 2023, pp. 12\,137--12\,146.

\bibitem{mathieu2015deep}
M.~Mathieu, C.~Couprie, and Y.~LeCun, ``Deep multi-scale video prediction beyond mean square error,'' \emph{arXiv preprint arXiv:1511.05440}, 2015.

\bibitem{shi2023video}
C.~Shi, C.~Sun, Y.~Wu, and Y.~Jia, ``Video anomaly detection via sequentially learning multiple pretext tasks,'' in \emph{Proceedings of the IEEE/CVF International Conference on Computer Vision}, 2023, pp. 10\,330--10\,340.

\bibitem{gong2019memorizing}
D.~Gong, L.~Liu, V.~Le, B.~Saha, M.~R. Mansour, S.~Venkatesh, and A.~v.~d. Hengel, ``Memorizing normality to detect anomaly: Memory-augmented deep autoencoder for unsupervised anomaly detection,'' in \emph{Proceedings of the IEEE/CVF International Conference on Computer Vision}, 2019, pp. 1705--1714.

\bibitem{morais2019learning}
R.~Morais, V.~Le, T.~Tran, B.~Saha, M.~Mansour, and S.~Venkatesh, ``Learning regularity in skeleton trajectories for anomaly detection in videos,'' in \emph{Proceedings of the IEEE/CVF conference on computer vision and pattern recognition}, 2019, pp. 11\,996--12\,004.

\bibitem{yan2023feature}
C.~Yan, S.~Zhang, Y.~Liu, G.~Pang, and W.~Wang, ``Feature prediction diffusion model for video anomaly detection,'' in \emph{Proceedings of the IEEE/CVF International Conference on Computer Vision}, 2023, pp. 5527--5537.

\bibitem{zhang2023exploiting}
C.~Zhang, G.~Li, Y.~Qi, S.~Wang, L.~Qing, Q.~Huang, and M.-H. Yang, ``Exploiting completeness and uncertainty of pseudo labels for weakly supervised video anomaly detection,'' in \emph{Proceedings of the IEEE/CVF Conference on Computer Vision and Pattern Recognition}, 2023, pp. 16\,271--16\,280.

\bibitem{fioresi2023ted}
J.~Fioresi, I.~R. Dave, and M.~Shah, ``Ted-spad: Temporal distinctiveness for self-supervised privacy-preservation for video anomaly detection,'' in \emph{Proceedings of the IEEE/CVF International Conference on Computer Vision}, 2023, pp. 13\,598--13\,609.

\bibitem{van2010software}
N.~Van~Eck and L.~Waltman, ``Software survey: Vosviewer, a computer program for bibliometric mapping,'' \emph{scientometrics}, vol.~84, no.~2, pp. 523--538, 2010.

\bibitem{awasthi2023anomaly}
A.~Awasthi, S.~Ly, J.~Nizam, S.~Zare, V.~Mehta, S.~Ahmed, K.~Shah, R.~Nemani, S.~Prasad, and H.~Van~Nguyen, ``Anomaly detection in satellite videos using diffusion models,'' \emph{arXiv preprint arXiv:2306.05376}, 2023.

\bibitem{batzner2024efficientad}
K.~Batzner, L.~Heckler, and R.~K{\"o}nig, ``Efficientad: Accurate visual anomaly detection at millisecond-level latencies,'' in \emph{Proceedings of the IEEE/CVF Winter Conference on Applications of Computer Vision}, 2024, pp. 128--138.

\bibitem{nie2024interleaving}
Y.~Nie, H.~Huang, C.~Long, Q.~Zhang, P.~Maji, and H.~Cai, ``Interleaving one-class and weakly-supervised models with adaptive thresholding for unsupervised video anomaly detection,'' \emph{arXiv preprint arXiv:2401.13551}, 2024.

\bibitem{zanella2023delving}
L.~Zanella, B.~Liberatori, W.~Menapace, F.~Poiesi, Y.~Wang, and E.~Ricci, ``Delving into clip latent space for video anomaly recognition,'' \emph{arXiv preprint arXiv:2310.02835}, 2023.

\bibitem{sethi2023video}
A.~Sethi, K.~Saini, and S.~M. Mididoddi, ``Video anomaly detection using gan,'' 2023.

\end{thebibliography}



\begin{IEEEbiography}[{\includegraphics[width=1in,height=1.25in,clip,keepaspectratio]{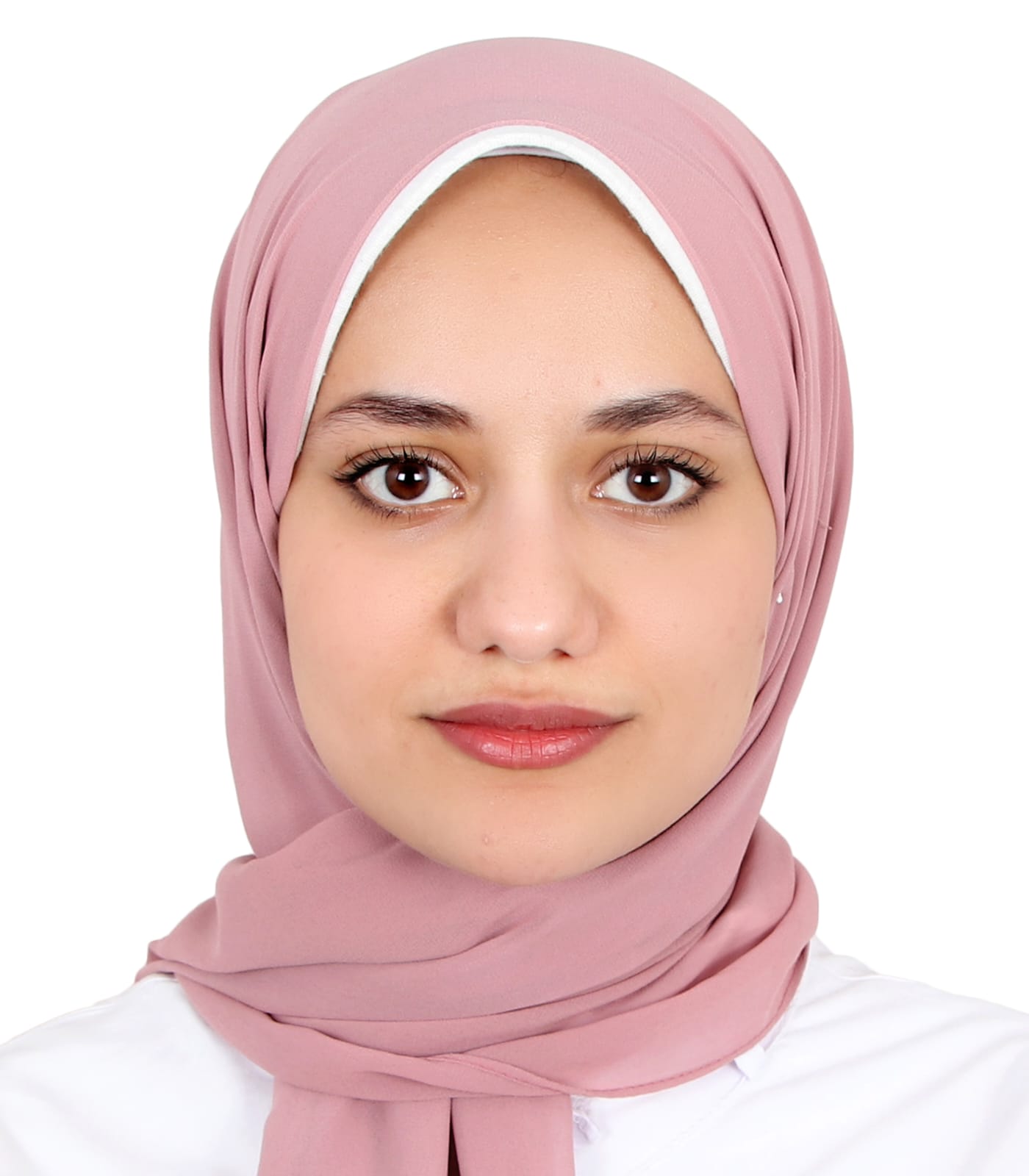}}]{Moshira Abdalla }
is a Graduate Research and Teaching Assistant (GRTA) and a Ph.D. in Electrical and Computer Engineering student at Khalifa University. She received a B.Sc. in Computer and Systems Engineering from Minya University, Egypt, in 2020 and her Master’s degree in Electrical and Computer Engineering from the University of Ottawa, Canada, in 2022. Her research is focused on Computer Vision, Anomaly Detection, and Artificial Intelligence (AI).
\end{IEEEbiography}



\begin{IEEEbiography}[{\includegraphics[width=1in,height=1.25in,clip,keepaspectratio]{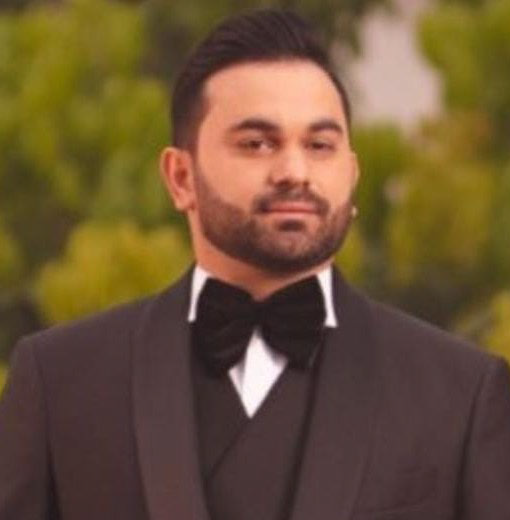}}]{Sajid Javed}
 is a faculty member at Khalifa University (KU), UAE. Prior to that, he was a research fellow at KU from 2019 to 2021 and at the University of Warwick, U.K, from 2017-2018. He received his B.Sc. degree in computer science from the University of Hertfordshire, U.K, in 2010. He completed his combined Master’s and Ph.D. degrees in computer science from Kyungpook National University, Republic of Korea, in 2017.

\end{IEEEbiography}

\begin{IEEEbiography}[{\includegraphics[width=1in,height=1.25in,clip,keepaspectratio]{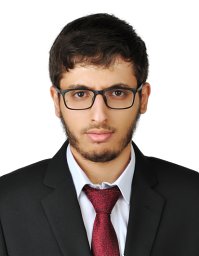}}]{ Muaz Al Radi}
 is a Graduate Research and Teaching Assistant (GRTA) and a Ph.D. in Electrical and Computer Engineering student at Khalifa University. He received his B.Sc. degree in Sustainable and Renewable Energy Engineering from the University of Sharjah, Sharjah, UAE, in 2020 and his M.Sc. in Electrical and Computer Engineering from Khalifa University, Abu Dhabi, UAE, in 2022. His research is focused on Computer Vision, Anomaly Detection, Vision-based Control, Artificial Intelligence (AI), and Robotics.

\end{IEEEbiography}


\begin{IEEEbiography}[{\includegraphics[width=1in,height=1.25in,clip,keepaspectratio]{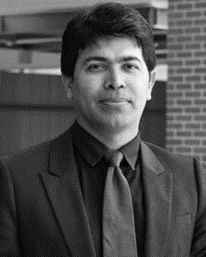}}]{  Anwaar Ulhaq }
received the Ph.D. degree in artificial intelligence from Monash University, Australia. He is currently working as a Senior Lecturer (AI) with the School of Computing, Mathematics, and Engineering, Charles Sturt University, Australia. He has developed national and international recognition in computer vision and image processing. His research has been featured 16 times in national and international news venues, including ABC News and IFIP (UNESCO). He is an Active Member of IEEE, ACS, and the Australian Academy of Sciences. As the Deputy Leader of the Machine Vision and Digital Health Research Group (MaViDH), he provides leadership in artificial intelligence research and leverages his leadership vision and strategy to promote AI research by mentoring junior researchers in AI and supervising HDR students devising plans to increase research impact.

\end{IEEEbiography}


\begin{IEEEbiography}[{\includegraphics[width=1in,height=1.25in,clip,keepaspectratio]{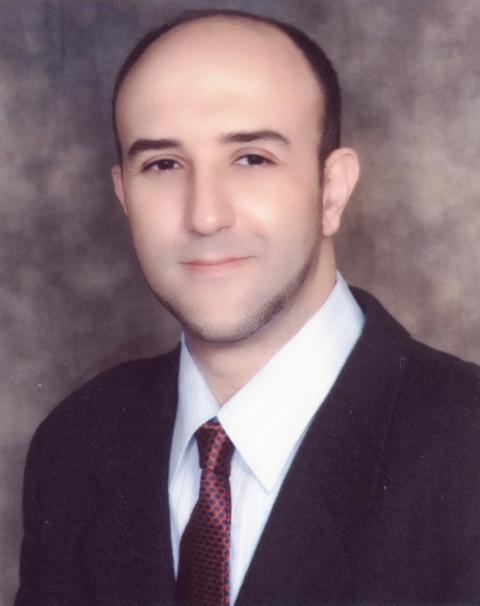}}]{ Naoufel Werghi}
 is a Professor at the Department of Computer Science at Khalifa University for Science and Technology, UAE. He received his Habilitation and PhD in Computer Vision from the University of Strasbourg. His main research area is 2D/3D image analysis and interpretation, where he has been leading several funded projects related to biometrics, medical imaging, remote sensing, and intelligent systems.

\end{IEEEbiography}

\EOD

\end{document}